%% file: main.tex
\documentclass[letterpaper]{article} 
\usepackage{aaai25}  
\usepackage{times}  
\usepackage{helvet}  
\usepackage{courier}  
\usepackage[hyphens]{url}  
\usepackage{graphicx} 
\usepackage{natbib}  
\usepackage{caption} 
\usepackage{algorithm}
\usepackage{algpseudocode}
\usepackage{newfloat}
\usepackage{listings}
\DeclareCaptionStyle{ruled}{labelfont=normalfont,labelsep=colon,strut=off} 
\floatstyle{ruled}
\newfloat{listing}{tb}{lst}{}
\floatname{listing}{Listing}
\usepackage{multirow}
\usepackage{graphicx}
\usepackage{amsfonts}
\usepackage{subcaption}
\usepackage{threeparttable}
\usepackage{booktabs}
\usepackage{longtable}
\usepackage{xcolor}
\usepackage{makecell}
\newcolumntype{L}{>{\centering\arraybackslash}m{3cm}}
\newcolumntype{M}{>{\arraybackslash}m{13cm}}
\newcount\myloopcounter
\newcommand{\repeatit}[2][10]{%
  \myloopcounter0
  \loop\ifnum\myloopcounter < #1 
  #2%
  \advance\myloopcounter by 1 %
  \repeat 
}

\usepackage{xspace} 
\newcommand{\tool}{EDCIM\xspace}
\newcommand{\edrepsilon}{0.1\xspace}

\usepackage{soul}
\usepackage{algorithm}
\usepackage{algpseudocode}
\usepackage{listings}
\definecolor{codebg}{rgb}{0.95,0.95,0.95}
\definecolor{codegray}{rgb}{0.5,0.5,0.5}
\definecolor{codepurple}{rgb}{0.58,0,0.82}
\definecolor{backcolour}{rgb}{0.95,0.95,0.92}

\def\conds{C}
\lstdefinestyle{mystyle}{
    backgroundcolor=\color{codebg},   
    commentstyle=\color{codegreen},
    keywordstyle=\color{magenta},
    numberstyle=\tiny \color{codegray},
    stringstyle=\color{codepurple},
    basicstyle=\ttfamily \scriptsize,
    breakatwhitespace=false,         
    breaklines=true,                 
    captionpos=b,                    
    keepspaces=true,                 
    numbers=left,                    
    numbersep=5pt,                  
    showspaces=false,                
    showstringspaces=false,
    showtabs=false,                  
    tabsize=2
}
\title{Error Detection and Correction for \\Interpretable Mathematics in Large Language Models}
\author{
    Yijin Yang\textsuperscript{\rm 1},
    Cristina Cornelio\textsuperscript{\rm 2},
    Mario Leiva\textsuperscript{\rm 3},
    Paulo Shakarian\textsuperscript{\rm 4}
    }
\affiliations{
    \textsuperscript{\rm 1}Arizona State University, Tempe, AZ \\
    \textsuperscript{\rm 2}Samsung AI, Cambridge, UK \\
    \textsuperscript{\rm 3}Universidad Nacional del Sur, Buenos Aires, Argentina \\
    \textsuperscript{\rm 4}Syracuse University, Syracuse, NY \\
    yyang491@asu.edu, c.cornelio@samsung.com, mario.leiva@cs.uns.edu.ar, pashakar@syr.edu}

\usepackage{bibentry}

\begin{document}

\maketitle

\input{sections/abstract}

\input{sections/introduction}

\input{sections/method}

\input{sections/exp-setup}

\input{sections/results}

\newpage
\section*{Acknowledgments}
This research was supported by DARPA cooperative agreement HR00112420370 (MCAI) and by Army Research Office (ARO) grant W911NF-24-1-0007.

\bibliography{mybib}

\newpage
~
\newpage
\appendix
\input{sections/appendix}

\end{document}

%% file: sections/abstract.tex
\begin{abstract}

Recent large language models (LLMs) have demonstrated the ability to perform explicit multi‑step reasoning such as chain‑of‑thought prompting. However, their intermediate steps often contain errors that can propagate leading to inaccurate final predictions. Additionally, LLMs still struggle with hallucinations and often fail to adhere to prescribed output formats, which is particularly problematic for tasks like generating mathematical expressions or source code.
%
This work introduces \tool ({\bf E}rror {\bf D}etection and {\bf C}orrection for {\bf I}nterpretable {\bf M}athematics), a method for detecting and correcting these errors in interpretable mathematics tasks, where the model must generate the exact functional form that explicitly solve the problem (expressed in natural language) rather than a black-box solution.
\tool uses LLMs to generate a system of equations for a given problem, followed by a symbolic error-detection framework that identifies errors and provides targeted feedback for LLM-based correction.
To optimize efficiency, \tool integrates lightweight, open-source LLMs with
more powerful proprietary models, balancing cost and accuracy.
This balance is controlled by a single hyperparameter, allowing users to control the trade-off based on their cost and accuracy requirements. 
Experimental results across different datasets show that \tool significantly reduces both computational and financial costs, while maintaining, and even improving, prediction accuracy when the balance is properly configured.
\end{abstract}

%% file: sections/introduction.tex
\section{Introduction}
\label{sec:Introduction}

Large language models (LLMs) have shown significant potential for converting unstructured natural language (NL) text into structured formats suitable for downstream tasks, such as equation solvers or code compilers.   This capability has been applied to a wide range of tasks, including solving math problems \cite{lewkowycz2022solving}, combinatorial optimization problems \cite{ye2023satlm,zhao2023investigating}, and declarative logic \cite{he2023solving}.  However, despite these advances, LLMs often produce intermediate steps with errors, which can propagate to the final output and lead to inaccurate predictions. This is particularly problematic for structured outputs like mathematical equations, where even minor errors can result in completely incorrect solutions.
Moreover, LLMs are prone to hallucinations and frequently struggle to adhere to strict output formats. To address these challenges, recent work has focused on re-prompting methods to reduce errors and improve the reliability of LLM outputs~\cite{gou2023critic,an2023learning}.  
However, these methods often require multiple re-prompting steps, as they do not have explicit error detection, resulting in significant resource consumption and high costs,  especially when using powerful state-of-the-art LLMs.


To address these challenges, we introduce \tool ({\bf E}rror {\bf D}etection and {\bf C}orrection for {\bf I}nterpretable {\bf M}athematics), a novel error detection and correction framework specifically designed for interpretable mathematical reasoning and inspired by concepts from meta-cognition (``thinking about thinking'')~\cite{flavell1976metacognitive,didolkar2024metacognitive}.
Unlike prior methods~\cite{gou2023critic}, which assume every response is potentially incorrect and thus re-query every sample, \tool uses a more selective approach. 
It combines lightweight, open-source LLMs for initial response generation with more powerful, cloud-based models for targeted error correction, reducing overall costs while maintaining high accuracy. 
Alternatively, it can also be used to enhance the performance of an LLM, whether local or cloud-based, by enabling it to iteratively correct and refine its own outputs.
This approach is made possible via the use of symbolic Error Detection Rules, adapted from \citet{kricheli2024error}, which identify specific error patterns in the generated equations before deciding whether to trigger a correction via re-prompting.
This approach also brings the benefit of explainability, as it provides explicit feedback on the potential mistakes made by the LLM.




A key feature of \tool is its ability to balance the cost and accuracy via a single hyperparameter,  which directly controls the re-prompt rate. This flexibility is fundamental for practical applications, allowing users to adapt to different computational budgets. 
For example, in large-scale experiments or industrial systems that involve millions of LLM queries, controlling cost and latency becomes as important as accuracy.
Unlike prior methods with limited re-prompting control, \tool offers a simple and effective way to manage this trade-off.
%
Our experiments show that using local models (Phi3) for initial solutions, combined with selective cloud-based (DeepSeek or GPT4o) queries for error correction, significantly reduces costs with minimal accuracy loss. 
Specifically, \tool re-prompts in only about one-third of cases while maintaining over $90\%$ of the full re-prompt accuracy. 
%
Another major contribution of our work is \tool's improved correction quality: even when full correction is  not possible, it consistently bring generated equations closer to the ground truth.
Finally, although \tool was designed for mathematical reasoning, our approach can be extended to other structured tasks, such as declarative logic, program synthesis, and combinatorial optimization.

LLMs are computationally intensive, consuming significant energy and contributing to the carbon footprint of AI systems. By providing a flexible framework that allows users to control the balance between computational cost and accuracy, \tool enables more efficient use of these models, reducing their environmental impact. Additionally, by improving the reliability of generated outputs through targeted error detection and correction, our approach enhances the safety of AI systems in applications where even small errors can lead to significant consequences.

\textbf{Limitations.}~ \tool performance depends on the selection of the error detection rules, which currently is done manually by a domain expert.  This can be particularly difficult when the task under consideration is very difficult or under-defined, where explicit error patterns are harder to define.  Moreover, the error detector require supervised data for learning which are the relevant rules. However, the number of samples required are minimal ($\sim$100). Additionally, the final solution relies on symbolic solvers, which are known to have scalability limitations.



\textbf{Related work.}~
Imposing constraints on neural networks and correcting their outputs is a challenging and widely studied topic in AI~\cite{cornelio_2023_NASR}. This is especially true for LLMs, where recent research has focused on correcting their outputs across many use cases~\cite{kamoi2024can,mishra2024correcting,pan2023automatically,upadhyaya2024internalized}.  
Most relevant to our work is CRITIC~\cite{gou2023critic}, a framework that, inspired by human reasoning, enables LLMs to self-correct by simulating a critique-and-revise loop with structured reasoning and feedback from external tools, such as calculators or code interpreters.
Importantly, CRITIC assumes that all the responses are incorrect and asks an LLM to revise its answers iteratively multiple times without distinguishing whether the answers are truly incorrect or not. 
Several recent methods have expanded upon or complemented the principles behind CRITIC. For instance, REFINER~\cite{paul2023refiner} introduced a similar self-refinement mechanism but focused on open-ended questions and factual generation, using internal consistency checks and a feedback model. Self-Refine~\cite{madaan2023self} adopts an edit-based feedback loop, where the model iteratively improves its answers based on automatically generated feedback.

Another relevant line of work is program-aided reasoning. While methods like PAL and MathCoder apply verification unconditionally, EDCIM, with its metacognitive EDR layer, evaluates initial outputs and selectively decides when to engage in costly correction, offering granular control over resource usage.

There has been a variety of approaches proposed to detect errors in LLM outputs, specifically focused on hallucinations~\cite{huang2023survey,su2024unsupervised,friel2023chainpoll}.  However, there has been limited work on detecting errors in the context of using LLMs to convert unstructured text into structured format for down-stream tasks. Works on such translation have noted high error rates involved in creating mathematical formulas~\cite{imani2023mathprompter,yamauchi2023lpml}, combinatorial problems~\cite{ye2023satlm}, and declarative logic programs~\cite{pan2023logic}\cite{gilpin2018explaining}.  However, none of these works has focused on specific mechanisms to detect errors.  Specific to the creation of mathematical formulae from unstructured text (focus of this paper) the work of \citet{shakarian2023independent} shows that certain characteristics of math word problems can be used to predict LLM errors with a simple classification model while the work of \citet{ngu2024diversity} estimates LLM uncertainty by computing metrics like entropy and Gini impurity over multiple responses to the same prompt.  
We build on these ideas to create a broader framework for both error detection and correction, allowing the system to learn which empirically observed error conditions are most relevant for a particular dataset. The work on symbolic error detection rules has been applied in a variety of tasks including image classification~\cite{kricheli2024error}, movement trajectory classification~\cite{xi2023rule} and time-series prediction~\cite{lee2024metal}. 
However, to our knowledge it has not been applied to the output of LLM's.


%% file: sections/method.tex
\section{The \tool Method}
\label{sec:Methods}

\begin{figure*}
    \centering
    \includegraphics[width=\linewidth]{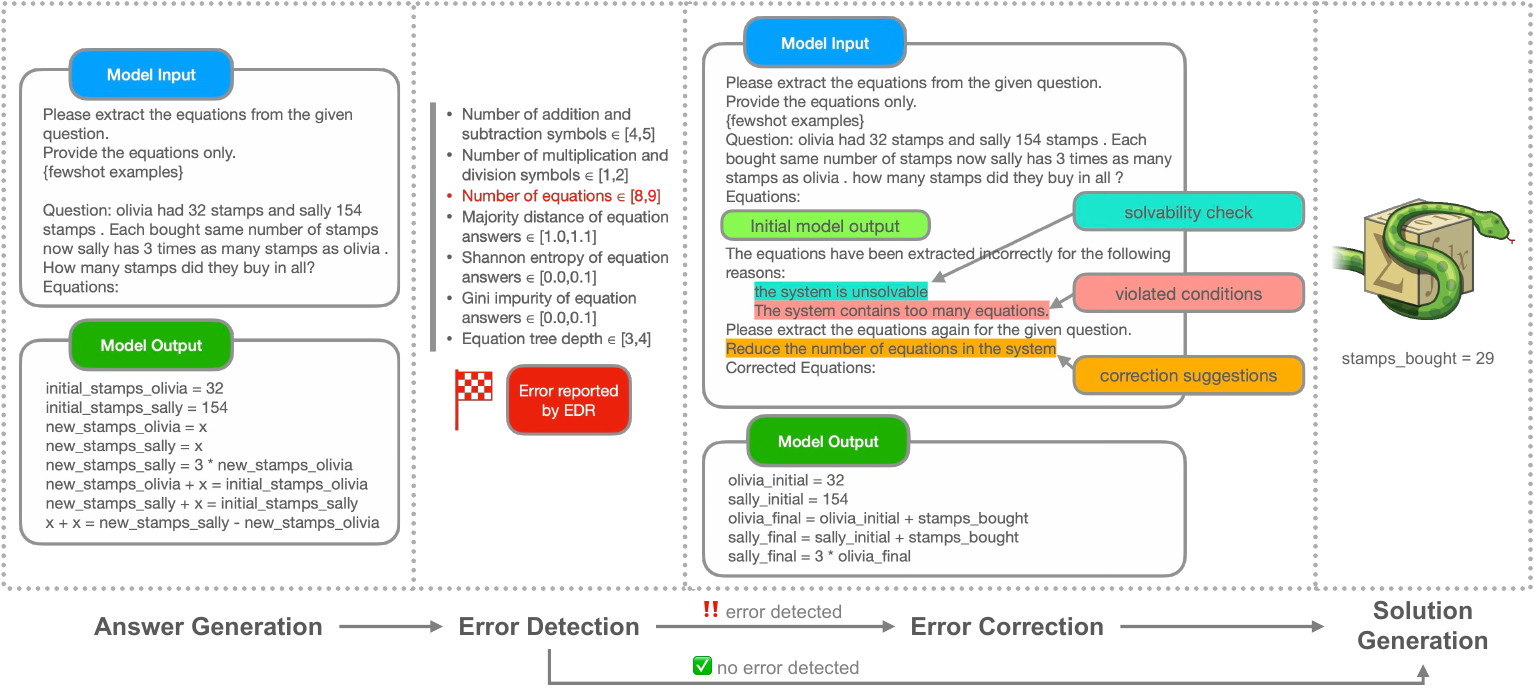}
    \caption{Overview of our system: (1) an LLM generates an initial system of equations from the problem description; (2) symbolic error detection identifies errors and produces justifications and guidance; (3) an LLM uses this information for context-aware error correction; and (4) the corrected equations are passed to a symbolic solver to generate the final numerical solution.}
    \label{fig:pipeline}
\end{figure*}

Our method, \tool ({\bf E}rror {\bf D}etection and {\bf C}orrection for {\bf I}nterpretable {\bf M}athematics), focuses on the task of solving mathematical problems by generating explicit systems of equations, which can then be used to compute the final answer via a symbolic solver. This task is significantly more challenging than simply generating the final numerical result, as it requires the model to produce interpretable outputs that humans can fully understand and verify.
Our system follows a 4-step process. First, an LLM generates an initial system of equations based on the natural language (NL) description of a mathematical problem provided as input. In the second step, a symbolic error detection model analyzes this initial output, identifying errors and producing insights for each detected one. These insights are then used in the third step, where an LLM leverages this feedback to produce a revised, corrected version of the equation system. Finally, the corrected equations are passed to a symbolic solver, which computes the final numerical solution. See Figure~\ref{fig:pipeline} for an overview.

More formally, \tool takes as input a natural language text $T$, which is first processed by an large language model, $\mathcal{LLM}_1$ to generate a set of equations, $X_1$ ($\mathcal{LLM}_1: T \mapsto X_1$). 
This initial system of equations is then analyzed by an error detection module, $EDR$, which identifies potential errors and produces a corresponding set of explanations and suggestions. 
$EDR$ output is then converted into natural language and concatenated in the text $J$ ($EDR_{NL}: X_1 \mapsto J$). 
In the next step, the justifications $J$ are concatenated to the original text $T$ and the initial equations $X_1$ forming the input for a second large language model, $\mathcal{LLM}_2$, which generates a revised set of equations, $X_2$ ($\mathcal{LLM}_2: (T,X_1,J)\mapsto X_2$). 
Finally, this corrected system of equations $X_2$ is passed to a symbolic solver to compute the final numerical solution, $S$ ($solve: X_2 \mapsto S$). 
The complete system can be summarized as: 
$ S = solve(~\mathcal{LLM}_2(T,~\mathcal{L}_1(T),~EDR_{NL}(\mathcal{LLM}_1(T))~) $.
In what follows, we present a detailed overview of each component in our framework, followed by the implementation details used in our experiments.

\subsection{Answer Generation}
Answer generation is the first step in our pipeline, where a large language model $\mathcal{LLM}_1$ is employed to transforms the initial, unstructured natural language text of the mathematical problem into a structured set of equations. 
The goal is to produce a complete system of equations $X_1$ that accurately captures the mathematical relationships described in the input text, as proposed in recent works for mathematical~\cite{gou2023critic} or logical reasoning~\cite{pan2023logic}.
An example of this process can be found in Figure~\ref{fig:pipeline}.
This is achieved by providing the LLM with carefully designed prompts (available in Appendix~\ref{app:prompts}) that include custom instructions and few-shot examples to guide the model towards the desired output format.
In particular, we instruct the LLM to produce equations in the same format required by the downstream symbolic solver, ensuring compatibility. 

To guide the LLM in generating outputs and to respect the required syntax, we incorporate a set of few-shot examples. These examples are manually curated to cover a broad range of mathematical word problem (MWP) types, inspired by external sources such as high school math textbooks.
Specifically, we reviewed several textbooks and noticed that MWPs are often organized into chapters based on problem types. Following this approach, we categorized the MWPs into classes and selected a representative example for each category, including: (1)~geometric;  (2)~numeric operations (addition, subtraction, multiplication, and division); (3)~monetary; (4)~chemistry solutions and mixtures; (5)~age; (6)~rate: speed, distance and time (e.g., $speed * time = distance$); and (7)~rate: work and time (e.g., $rate * time = work\_done$).
The full list of examples is available in Appendix~\ref{app:few_shot}.
%
%
However, despite providing explicit syntax guidelines and curated few-shot examples, the outputs generated by the LLM  might sill not always match the required format. To address this, we employ a parser to clean, simplify, standardize, and interpret the generated equations, ensuring they are properly formatted to match the expected solver syntax.

\subsection{Error Detection}
In the second step, we leverage 
EDR~\cite{kricheli2024error} to evaluate the system of equations $X_1$ generated by the LLM and identify potential errors. EDR (Metacognitive Error Detection Rules) is a neuro-symbolic approach that combines human-designed rules with statistical supervision for error detection. 
EDR framework take a set of predefined error detection rules, and if the preconditions of one or more of these rules are satisfied, the generated equations are flagged as containing errors.
The specific rules that are relevant for a given dataset are identified in advance during training, starting from a manually defined set of candidates. The learning process is described in what follows.


Using the first-order logic notation introduced by \citet{xi2023rule} and by \citet{kricheli2024error}, we denote one or more LLM responses to a prompt for a given query as $X$.  We say some condition (denoted $\mathit{cond} \in C$) is true for sample $X$ as $\mathit{cond}(X)$.  In the domain of generating mathematical equations, an example condition for a single sample could be the number of addition operations exceeding a certain amount.  An example where $X$ is a set of LLM responses (to the same prompt) is that the entropy of the set of responses falls within a certain range.  We denote as $C'$ the subset of conditions $C$ that are known to cause errors (e.g., identified by analyzing responses generated from a set of prompts sent to the LLM, where the ground truth is known). Based on this subset, we can specify a set of error detection rules as follows:
$ \mathit{cond}\in \mathit{C'}:\mathit{error}(X) \leftarrow \mathit{cond}(X)$.

Using ideas from the literature~\cite{ngu2024diversity, shakarian2023independent}, we first define the set of candidate conditions $C$ that capture potential error patterns. We then identify a subset of conditions $C'$ most likely to cause errors on a specific dataset.
This subset is learned by evaluating the relevance of each candidate condition on a training set.
A key feature of this approach is that when a rule executes on a given LLM response, we have precise knowledge of which conditions were present. This information is later leveraged to develop custom prompts that enable fine-grained and informed LLM-based correction.

Algorithm~\ref{alg:ruleSelct} reviews the \textsf{DetRuleLearn} rule-learner of \citet{xi2023rule}, adapted for our use case. Given $N$ training samples, for a subset $C'\subset C$, let $POS_{C'}$ denote the number of samples with at least one condition in $C'$ causes the LLM to produce an error, and $NEG_{C'} = N-POS_{C'}$. The Recall Reduction Threshold hyperparameter, $\varepsilon \in (0,1]$, is a critical parameter for error detection as it controls the balance between detection coverage and precision.
A higher value of $\varepsilon$ allows for a more aggressive detection strategy, resulting in a larger number of detected errors but potentially increasing the rate of false positives. In contrast, a lower $\varepsilon$ value prioritizes precision, reducing the rate of false positives but potentially missing some errors.
\textsf{DetRuleLearn}  leverages ideas from constrained submodular optimization and we refer the reader to \citet{xi2023rule} and \citet{shakarian2025probabilistic} for a formal treatment of this algorithm in a classification setting.

\begin{algorithm}[h] 
\caption{\textsf{DetRuleLearn}~\cite{xi2023rule}}
\label{alg:ruleSelct}
\begin{algorithmic}
    \State {\bfseries Input:} Recall reduction threshold $\varepsilon$, Condition set $\conds$
    \State {\bfseries Output:} Subset of conditions $C'$
    \State{$C':=\emptyset$}
    \State{$C^* := \{ c \in \conds \textit{ s.t. } NEG_{\{c\}} \leq \varepsilon \cdot N$ \} }
    \While{$C^* \neq \emptyset$}
        \State{$c_{best}=\arg\max_{c \in C^*} POS_{C'\cup\{c\}}$}
        \State{Add $c_{best}$ to $DC_i$}
        \State{$C^* :=\{ c \in \conds\setminus C'\textit{ s.t. }  NEG_{C'\cup\{c\}} \leq \varepsilon \cdot N$ \} }
    \EndWhile
    \State{\textbf{return} $C'$}
\end{algorithmic}
\end{algorithm}

\textbf{Error correction.}~  
A key component of our approach is the ability to generate custom, context-aware prompts based on detected errors, using a set of interpretable Error Detection Rules (EDRs). These rules not only identify potential failure modes in the output of the LLM (e.g., overuse of mathematical symbols, incorrect structure, too much diversity among multiple queries) but also provide instructional feedback that can be directly translated into new prompts. 
The correction process begins by constructing a new prompt, which is formed by concatenating the original query $T$, the initial solution set $X_1$, and the insights provided by the triggered error detection rules. This prompt is then passed to an LLM, allowing it to generate a revised solution while being informed about the specific causes of the detected errors in a fine-grain manner.

The insights provided by EDR are converted in two main natural language components: violated conditions and recovery suggestions. 
Violated conditions describe the error detection rules that were triggered during the evaluation of the initial solution. These are converted in natural language by using predefined templates that we manually created when defining the set $C$ of candidate rules for the EDC framework.
Recovery suggestions provide targeted guidance for correcting the identified errors. These, also defined using natural language templates, are created during the initial definition of the candidate detection rules. For example, if the error is "The system contains too many equations", the corresponding recovery suggestion might be "Reduce the number of equations in the system". 
Like the violated conditions, these templates are manually prepared, but it is important to note that they could also be automatically generated using an LLM.
In addition to the recovery suggestions, we include a fixed one-shot example from the training set, which shows how to correct errors based on the triggered rules for a specific instance. The corrections for this example were manually written, and cover only a subset of the full set of conditions considered in our framework.





% ...


% ...

% ...
The final correction prompt is then constructed by concatenating, in the following order, the original query $T$, the initial system of equations $X_1$, the violated conditions block and the recovery suggestions including the few-shot example. This prompt is then used to guide the LLM in producing the corrected solution $X_2$. As with the answer generation step, the corrected output is then parsed to ensure its consistency to the required syntax.
In Figure \ref{fig:pipeline} we show an example of the re-prompt for our running example, where the dynamic template components are marked by different colors. 
It is important to note that \tool only attempts to apply a correction if an error is detected. If no errors are identified, the original system of equations remains unchanged, resulting in $X_1=X_2$.


\subsection{Solution Generation} 
In the final step of our framework, we rely on an external solver that takes in input the corrected system of equations $X_2$, produced by $\mathcal{LLM}_2$, and output the final numerical solution $S$. The symbolic solver guarantees a correct solution to the input system, provided that the system is well-defined and consistent.
In cases where the input system is unsolvable, the solver returns an empty set, indicating no possible solutions. If the system instead is ill defined (e.g., syntax errors) the solver will produce an error. Both of this cases will be tagged as incorrect solutions.
If the system of equations is under-determined, the solver identifies multiple valid solutions. In this case, for evaluation purposes, we check if the ground-truth solution appears in this set.


\subsection{Local Open-Source LLMs vs Cloud-Based Proprietary LLMs.} 
Our framework combines the advantages of lightweight, open-source LLMs running locally with the superior performance of larger, proprietary models accessed via API. The goal is to have an effective balance between computational cost, financial efficiency, and predictive accuracy.
We refer to small, open-source models deployed on local hardware with low computational requirements as \textit{local LLMs}. These models offer zero additional inference costs and offline accessibility, though their performance is typically limited by smaller model sizes and constrained computational resources.
%
In contrast, we refer to powerful, proprietary models hosted on commercial platforms as \textit{cloud-based LLMs}. 
These models leverage extensive computational resources, providing superior reasoning and problem-solving capabilities. However, they come with substantial usage costs, making them less cost-effective for high-volume tasks (such as big datasets).

The goal of our approach is to optimally combine these two types of LLMs to maximize accuracy while minimizing cost. 
The key setting that we considered is where a local LLM is used to generate an initial, approximate solution. 
If this initial output is correct, it is accepted, reducing computational and financial costs, but if errors are detected, it is refined using a more powerful cloud-based LLM.
%
Another relevant setting is where we enhance the performance of an LLM, whether local or cloud-based, by enabling it to iteratively correct and refine its own outputs.

%% file: sections/exp-setup.tex
\section{Experimental set-up}
\label{sec:set-up}

For our experiments, we utilize different LLM models: For the local-based, we use the \texttt{Phi-3 Mini 128K} model~\cite{abdin2024phi}, a compact, open-source LLM specifically optimized for reasoning tasks; For the cloud-based models, we submit API requests to \texttt{GPT4o} (OpenAI)~\cite{achiam2023gpt}  and \texttt{DeepSeek-V2-R1} (DeepSeek)~\cite{liu2024deepseek}, both of which are larger, more powerful models known for their advanced reasoning capabilities.
The answer generation is performed 10 times for each sample to evaluate the diversity among responses. 
The answer correction is only called if an error has been identified and the LLM is only queried once for each sample. 
All local model inference experiments were conducted on an NVIDIA A100 GPU.

We use SymPy~\cite{sympy} as our symbolic solver, a python library for symbolic mathematics that we use to solve the mathematical equations generated by the LLMs. It also serves as the backbone for our custom parser, which attempts to fix common syntax errors to correctly load the equations. This involves tasks like equation manipulation and simplification. While our parser handles the majority of cases, it is not 100\% accurate, meaning that a small number of outputs may still be incorrectly formatted and thus classified as errors.



In the generation prompt, we instruct the LLM to act as a math assistant that converts word problems into raw mathematical equations. The prompt specifies that the output should consist solely of the equations, without any explanations, labels, or additional text, with each equation on a separate line. Variable names should be limited to letters and underscores, and only SymPy-compatible mathematical operators ($+$, $-$, $*$, $/$, $=$) are permitted (e.g., $age\_sarah = 2 * age\_brother ~ \backslash n ~ age\_sarah + age\_brother = 27$). The full prompts are provided in Appendix~\ref{app:prompts}.



\begin{figure}[h]
  \centering
  \begin{subfigure}{0.8\linewidth}
    \includegraphics[width=\linewidth]{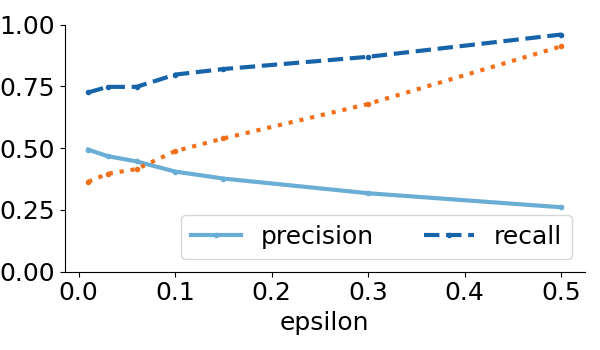}
    \caption{DRAW-1k}
  \end{subfigure}
  \hspace{1cm}
  \begin{subfigure}{0.8\linewidth}
    \includegraphics[width=\linewidth]{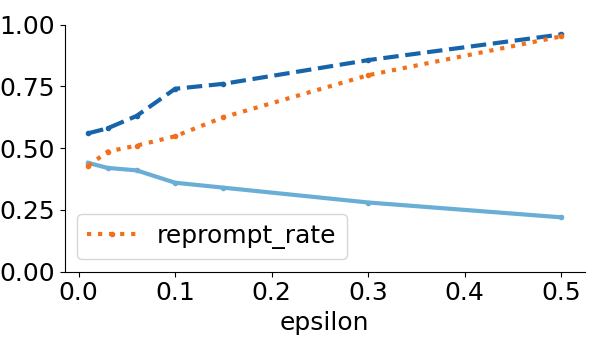}
    \caption{GSM-8k}
  \end{subfigure}
  \caption{Effect of $\varepsilon$ on detection precision, recall and re-prompt rate. Higher $\varepsilon$ values increase \tool's detector recall but reduce precision. Zero re-prompting means only local correction. Results are shown for DRAW-1K (a) and GSM-8K (b) using Phi-3 as generator and GPT-4o as corrector.}
  \label{fig:epsilon}
\end{figure}
\subsection{Detection Rules} 
The set $C$ of candidate conditions is defined starting from a collection of general rule categories, or meta-rules, which capture common error patterns identified in the literature~\cite{xi2023rule}. 
Each condition within the set $C$ is a specific grounding, or parameterization, of a meta-rule, created by assigning numerical values to the intervals boundaries that define that rule.
For example, the meta-rule ``Number of equations $\in [a,b]$'' can be grounded in several conditions in $C$ such as ``Number of equations $\in [4,5]$'' or ``Number of equations $\in [8,9]$''.


In this work we consider two primary categories of meta-rules, identified based on empirical observations and domain knowledge. 
(1) \textit{Equation Complexity Measures} describe structural aspects of the equations, such as the depth of the equation parsing tree and the number of the operators. More complex equations, with a deeper tree or with a higher number of mathematical symbols, correspond to more difficult questions that thus carry a higher likelihood of containing an error.
In this category we define different types of meta-rules, including the average depth of the equations parsing tree, the number of additions and subtractions, the number of multiplications and divisions, and the total number of equations.
(2) \textit{Diversity Measures} analyze the response consistency when querying LLMs multiple times on the same problem. 
If a model produces highly variable outputs across multiple queries on the same input problem, it indicates lack of confidence in its responses, increasing the likelihood of errors.
In this category we have different 
types of meta-rules, such as Shannon Entropy (a measure of the variability in the LLM's responses),  Gini Impurity (a measure of the likelihood of incorrect classification within a set of responses) and Jaccard distance to the core set of solutions 
(i.e., the Jaccard distance between the solutions of each system of equations and the largest common core solution set for individual variables, computed across multiple responses). 


Each category defines multiple fine-grained rules by grounding of its intervals parameters (as mentioned above), forming the pool $C$ of candidate detection rules.
During training, EDR error detector learns the optimal parameter intervals for each meta-rule, selecting the ranges that most effectively distinguish correct from incorrect outputs.
Thus, from the pool of candidate conditions $C$, a subset $C'$ is learned, containing the conditions most likely to indicate errors for a specific dataset. This learnt configuration is then used to classify responses in the test set.

\subsection{Datasets and Baselines}
We conduct our experiments on two widely used mathematical benchmark datasets: DRAW-1K~\cite{upadhyay2016annotating} and GSM-8K~\cite{cobbe2021gsm8k}. 
DRAW-1K (DiveRse Algebra Word problems) contains $1000$ algebra word problems crawled from \url{algebra.com}. Each problem is annotated with both the ground-truth formulae leading to the final numerical solution and the solution itself.
GSM-8K (Grade School Math problems)  is a dataset of $8500$ math word problems written by human authors and released by OpenAI. Each problem is accompanied by a detailed solution that includes the intermediate reasoning steps needed to reach the final answer, from which both the ground-truth equations and the final numerical solution can be recovered. 
%
Given that our error detection method, since symbolic, does not rely on large training sets, we modified the official train-test splits 
into a 1:9 train-test ratio, which we found to be the most effective configuration based on preliminary experiments (see Appendix~\ref{app:datasets} for more details). While state-of-the-art models achieve high performance on these benchmarks, our focus is to validate a cost-saving, controllable correction methodology.

We compare \tool against different methods:
(1) \textbf{LLM only}: running an LLM one time per sample;
(2) \textbf{SC}: running a LLM multiple times (10 times in our experiments) and adopting self-consistency (SC) which aggregates the multiple answers via majority voting;
(3) \textbf{SC+Solv.}: a refined variant of SC that performs majority voting only among solvable answers; and
(4) \textbf{CRITIC}~\cite{gou2023critic}: a framework that self-correct an LLM output by performing one or more LLM-critique-and-revise loops with feedback from external tools (e.g., python interpreters). In our experiments we re-prompt each sample once.

%% file: sections/results.tex
\begin{figure}[h]
  \centering
  \begin{subfigure}{0.8\linewidth}
    \includegraphics[width=\linewidth]{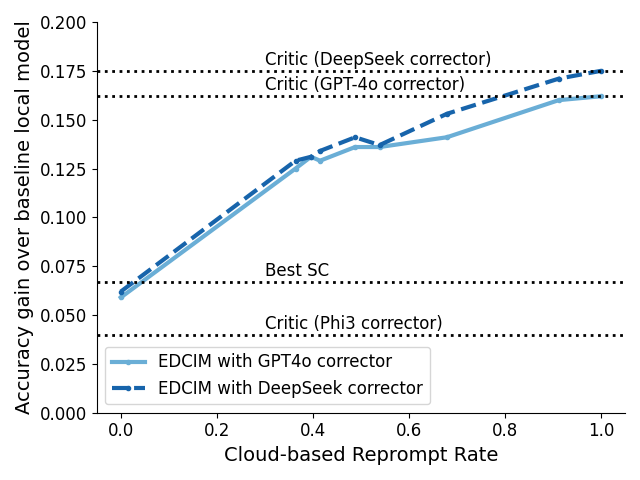}
    \caption{DRAW-1k}
  \end{subfigure}
  \begin{subfigure}{0.8\linewidth}
    \includegraphics[width=\linewidth]{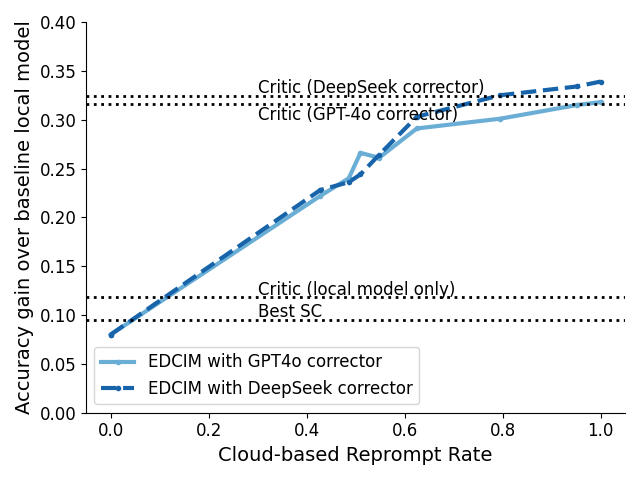}
    \caption{GSM-8k}
  \end{subfigure}
  \caption{Trade-off between accuracy gain over the baseline local model Phi-3 (answer generator) and re-prompt rate, as cloud-based re-prompt rate (of the error corrector model) varies (controlled by $\varepsilon$). Results are shown for DRAW-1K dataset (a), and GSM-8K dataset (b).}
  \label{fig:results}
\end{figure}

\section{Experimental results}
\label{sec:Experiments}


Our experimental results can be summarized as follows:
(1) Using local models for initial solutions and selective cloud-based queries for error correction significantly reduces costs, with only a small trade-off in accuracy: \tool re-prompts only about one-third of cases, unlike state-of-the-art that re-query every sample;
(2) \tool provides flexibility by allowing users to adjust the trade-off between cost and performance through a single hyperparameter $\varepsilon$, providing full control over this balance;
(3) Even when the corrector LLM is unable to fully correct the system of equations, it still improves the overall quality, bringing the equations closer to the ground truth.

\begin{table*}[h]
\caption{Comparison of our method \tool with baselines (LLM only, SC and SC-solvable) and the state-of-the-art CRITIC~\cite{gou2023critic} framework using all considered LLMs (Phi-3, GPT4o, and DeepSeek) and optimal value of $\varepsilon$. Reported metrics include Accuracy (ACC) and cloud-based re-prompt rate (percentage). \tool achieves the best trade-off between these two metrics.
}
\label{tab:results}
\centering
\small
\begin{tabular}{lllccccc}\toprule 
 & \multirow{2}{*}[0pt]{Answer} & \multirow{2}{*}[0pt]{Error}& \multicolumn{2}{c}{DRAW-1k} & &  \multicolumn{2}{c}{GSM-8k}
\\
\cmidrule{4-5} 
\cmidrule{7-8}
Method &Generator&Corrector& ACC  & re-prompt \%  &  & ACC  & re-prompt \% \\
\midrule
LLM only & GPT4o & -& 91.6 & - & & 84.4 & - \\
LLM only & DeepSeek& - & 92.4 & -  & & 86.2 & - \\
LLM only & Phi3 & - & 75.2 & -  & & 52.2 & - \\
SC & Phi3 & - &  
78.8 & - &  & 
60.4 & -\\
SC+Solv.&Phi3&-&  
81.9 & - &  & 
61.7 & -\\\midrule

\multirow{3}{*}[0pt]{CRITIC}&Phi3&Phi3&  
79.2 & - & &  
64 & -\\
&Phi3 & GPT4o&  
91.4& 100 &  & 
83.8 & 100\\
&Phi3 & DeepSeek&  
92.7 & 100 &  & 
84.6 & 100\\

\midrule

\multirow{3}{*}[0pt]{\tool}&Phi3&Phi3&  
78.8 & - &  & 
66.2 & -\\
&Phi3 &GPT4o&  
85.7 & 36 &  & 
74.4 & 43\\
&Phi3 &DeepSeek&  
87.8 & 36 &  & 
75.0 & 43\\

\bottomrule
\end{tabular}
\end{table*}

\subsection{Effect of hyperparameter selection for re-prompt rate control}\label{sec:epsilon}

Figure \ref{fig:epsilon} illustrates how the $\varepsilon$ hyperparameter in \tool's detector affects precision, recall, and re-prompt rate, using Phi-3 as answer generator and GPT4o as error corrector.
This parameter controls the  sensitivity of error detection, creating a trade-off where higher $\varepsilon$ values increase recall by detecting more errors, but at the cost of reduced precision due to more false positives. In contrast, lower $\varepsilon$ values improve precision but reduce recall, as more errors go undetected. 
The hyperparameter $\varepsilon$ provides flexibility by allowing users to adjust the trade-off between cost and performance. 
This is achieved by indirectly controlling the re-prompt rate, as higher $\varepsilon$ values increase the likelihood of re-prompting, leading to higher accuracy at a greater cost. Moreover, as shown in Figure~\ref{fig:epsilon} the relationship between $\varepsilon$ and the re-prompt rate is approximately linear, making this control highly predictable. Generally, directly controlling the re-prompt rate is challenging, and existing state-of-the-art systems lack effective mechanisms for this. In contrast, our approach, which uses EDC error detector to indirectly manage the re-prompt rate via the $\varepsilon$ parameter, offers a the ability to balance between computational expense and performance.

Figure~\ref{fig:results} shows the comparative performance of \tool, the baselines, and CRITIC against the local model (Phi-3) on DRAW-1K and GSM-8K datasets, as the re-prompt rates vary (indirectly with $\varepsilon$). 
It shows the trade-off between comparative accuracy (calculated as the difference between the ACC of each model and the ACC of the local model, Phi-3) and the average number of cloud-based queries.
We achieve the different re-prompt rates by sampling the values of $\varepsilon$ by grid search in $(0,0.5]$ with more density in low values (and manually adding $100\%$ re-prompt rate). 
From this figure, we can observe that as the re-prompt rate increases, the comparative accuracy shows a linear improvement. Notably, at a 100\% re-prompt rate, our method slightly outperforms CRITIC. 

Given the results of these two experiments we choose $\varepsilon = \edrepsilon$ as optimal value for both DRAW-1K and GSM-8K, as it has a good balance between precision and recall across these two datasets. Further improvements can be obtained by fine-tuning $\varepsilon$ value separately for each datasets.


\subsection{System performance at $\varepsilon = \edrepsilon$}

Table \ref{tab:results} reports the comparison, in terms of Accuracy (ACC) and cloud-based re-prompt rate, of \tool with the baselines and the state-of-the-art CRITIC framework. We used all considered LLMs (Phi-3, GPT4o, and DeepSeek) and the optimal value of $\varepsilon$ defined in Section~\ref{sec:epsilon}.  Our results indicate that \tool selectively re-prompts only about one third of cases, significantly reducing reliance on expensive cloud-based queries when compared to CRITIC, which re-prompts each sample. 
Re-prompting in 100\% cases achieves high accuracy but at substantial computational and financial cost, while using only local models avoids cloud-based queries but results in lower accuracy. \tool optimally balances local and cloud-based queries, extracting maximal value from cost-free local queries before resorting to cloud-based corrections. Overall, \tool demonstrates superior cost efficiency while maintaining competitive accuracy.
Moreover, on the GSM-8k dataset, \tool outperforms the state-of-the-art when only the local Phi-3 model is used for both answer generation and correction.


\begin{figure}[h]
  \centering
  \begin{subfigure}{\linewidth}
  \centering
    \includegraphics[width=\linewidth]{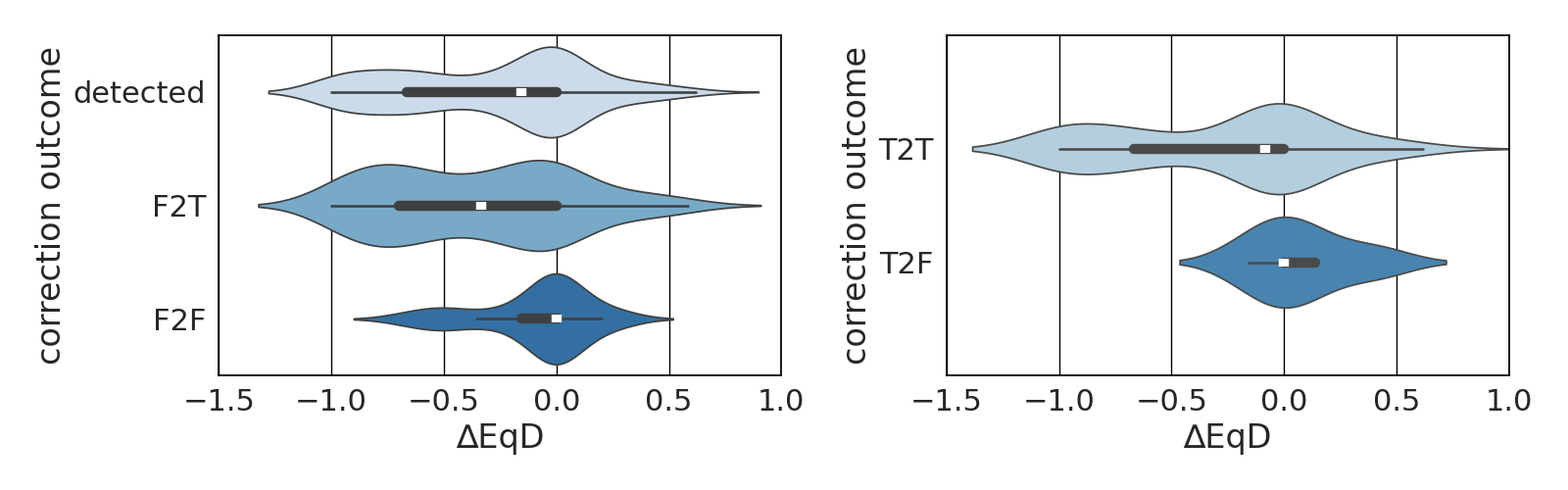}
  \end{subfigure}
  \caption{Change in Equation Distance $\Delta EqD$, before and after correction on DRAW-1K. The middle bar represents the median. Negative values indicate equations moved closer to the ground truth. Results use Phi-3 for answer generator and GPT4o for error correction and $\varepsilon = \edrepsilon$.  
  }
  \label{fig:violin}
\end{figure}

\subsection{Solution improvement analysis}
\label{subsec:sol-improvement-analysis}

We analyze the extent to which EDCIM can improve the quality of the generated equations, even in cases where it fails to produce a fully correct result.
To evaluate the quality of the generated equations we measure the distance between the ground truth equations and both the original and corrected equations.
We define a distance metric, \textit{Equation Distance} $EqD(X,Y)$ between two system of equations $X$ and $Y$ which is computed as the average of the following three metrics: 
(1) The distance between their numerical solutions;
(2) The structural distance, measured as $1-\frac{|N_1-N_2|}{\max{N1,N2}}$ where $N_1$ and $N_2$ are the total number of nodes in the parsing trees of the two systems; and 
(3) The complexity distance, calculated as  $\frac{|O_1-O_2|}{\max{O1,O2}}$ where $O_1$ and $O_2$ are the number of operators in the two systems.

We then evaluate the improvement in distance to the ground truth system of equations, defined as the difference $\Delta EqD = EqD(X_2,X^*) - EqD(X_1,X^*)$ where $X_1$ is the initial set of equations, $X_2$ is the corrected set (as defined in Section~\ref{sec:Methods}), and $X^*$ is the ground truth. Negative values indicate that the correction step moved the equations closer to the ground truth, while positive values indicate the opposite.

Figure~\ref{fig:violin} shows the values of $\Delta EqD$ when using Phi-3 as answer generator and GPT4o as error corrector on DRAW-1K dataset. The results show that re-prompting generally produces equations closer to the ground truth. For this analysis, we categorize the correction outcomes as follows: 
(1) \textbf{F2T (False-to-True)}, where the initial response was incorrect but corrected to the correct solution; 
(2) \textbf{T2F (True-to-False)}, where a correct response was incorrectly modified; 
(3) \textbf{T2T (True-to-True)}, where the response remained correct; 
(4) \textbf{F2F (False-to-False)}, where the response remained incorrect; and 
(5) all generated equations flagged as errors by \tool's detector, regardless of their correctness.

We observed significant improvement in the quality of solutions in all detections, with the bulk of improvement occurring in successfully corrected errors (F2T).  However, we also noted improvement in incorrect detections cases (T2T) where results from EDCIM were also found to be closer to the ground truth, likely refining equation structure without altering correctness.  
F2F cases are particularly interesting; while they remain incorrect, our analysis shows that the correction step more often improves the solution quality than degrades it. This suggests that even unsuccessful corrections can provide incremental value, reinforcing the overall effectiveness of our framework.
We note that false-positive detections improperly corrected (T2F, carrying the risk of over-correction) were infrequent in the analysis. Overall, these results confirm the effectiveness of our correction framework in reducing errors, even if only partially. To validate this finding, we conducted an study using several alternative methods to measure equation distance, including multiple graph embedding techniques and vector distance functions. A detailed presentation of this is available in Appendix~\ref{sec:equation-distance-analysis}.



\subsection{Ablation studies}


We conduct a series of ablation studies to evaluate the actual impact of different components in our framework. Specifically, we analyze the effects of various detector models, rule types, and prompt designs on \tool performance. For details, refer to Appendix~\ref{sec:Ablation}.

\noindent
\textbf{Impact of Error Detector Type.}~
We evaluated the impact of different error detection models, including full re-prompting, ground truth-based correction, equations solvability checks, and EDR without and with equations solvability checks.  Our results indicate that integrating both learned error detection and solvability checks, as used in \tool, provides the most accurate and comprehensive correction, reducing false negatives and capturing a broader range of error types.

\noindent
\textbf{Impact of Prompt Type.}~
We modified the prompt by removing prompt components (e.g., few-shot examples, correction suggestions, etc.) resulting in different settings, including the standard method employed in \tool and static prompts that do not take into account dynamic, example-dependent cues. The results indicate that the standard setting used in \tool provides the highest overall performance; violated conditions and correction suggestions lead to very similar additional information and that static prompts leads to the worst performance overall, highlighting the importance of targeted example-dependent feedback.

\noindent
\textbf{Impact of Rule Type.}~
We excluded some rules meta-rules category (see Section~\ref{sec:set-up}) from the set $C$ to see if they are both impacting error detection accuracy.
The results show that using both rule types increases the number of detected errors, resulting in higher re-prompt rates and overall accuracy, despite a slight reduction in precision.

For more details, refer to Appendix~\ref{sec:Ablation}.

\section{Final remarks and future work discussion}
To conclude, in this work, we introduced \tool, an error detection and correction framework for LLMs that uses a symbolic-based and interpretable error detection mechanism(EDRs). Unlike prior work, \tool selectively triggers corrections, allowing users to control the balance between cost and accuracy through a single hyperparameter. Our results show that this approach significantly reduces costs, with only a small trade-off in accuracy, and even partial corrections tend to improve the quality of the generated solutions. 
%
While our experiments focused on math problems, this approach is broadly applicable to other structured outputs such as logic, combinatorial optimization, and program synthesis. Future directions include automatic EDR learning, deeper integration with symbolic solvers, and extending the framework to new tasks like theorem proving and complex knowledge extraction. We also plan to assess performance on more challenging benchmarks, such as the MATH dataset, with the latest generation of LLMs. Notably, our initial pilot experiments suggest EDCIM can also improve state-of-the-art models (e.g., DeepSeek and GPT-4o), a direction best explored on these more complex datasets where performance is not already near-saturated.

%% file: sections/appendix.tex
\onecolumn

\section{Few-shot examples}\label{app:few_shot}

\subsection{Answer generation}
\begin{figure}[h]
    \centering
    \includegraphics[width=\linewidth]{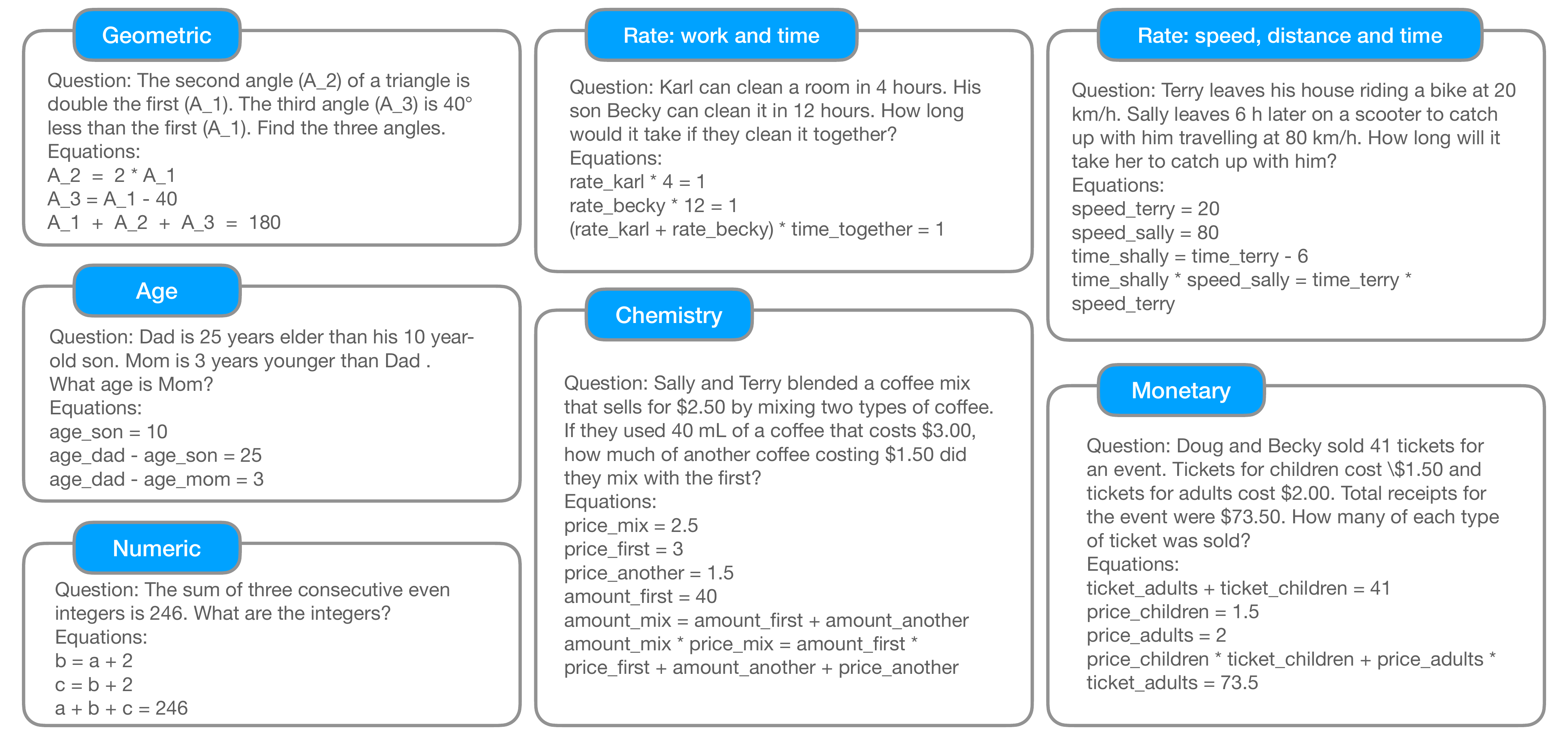}
    \caption{Few-shot examples for Answer Generation}
    \label{fig:fewshot-initial}
\end{figure}

Figure~\ref{fig:fewshot-initial} contains the  few-shot examples used as part of the input for the LLM generating the initial set of equations. As described in Section~\ref{sec:Methods} we used 7 different example categories: (1)~geometric;  (2)~numeric operations (addition, subtraction, multiplication, and division); (3)~monetary; (4)~chemistry (solutions and mixtures); (5)~age; (6)~rate: speed, distance and time; and (7)~rate: work and time.

\subsection{Answer correction}
\begin{figure}[h]
    \centering
    \includegraphics[width=0.53\linewidth]{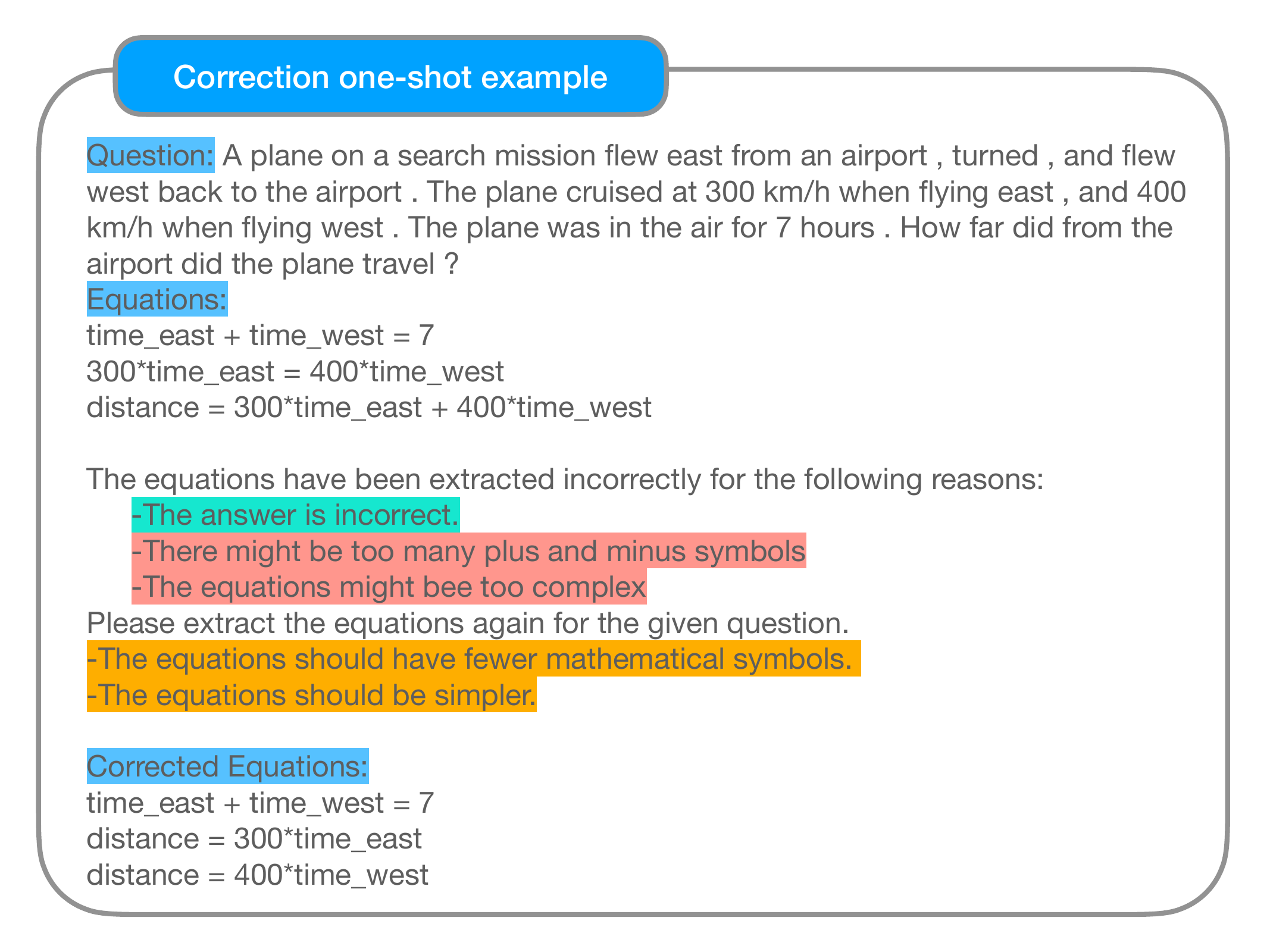}
    \caption{One-shot example for Answer Correction}
    \label{fig:fewshot-reprompt}
\end{figure}

In Figure~\ref{fig:fewshot-reprompt} is reported the one-shot example used as part of the input for the LLM performing the error correction.

\newpage
\section{Prompts templates}\label{app:prompts}

\subsection{System instruction}
\begin{lstlisting}
You are a math word problem solver. Given a math word problem, you need to extract the relevant information and convert it into a mathematical equation.
Your task is to identify the variables and the mathematical operations involved in the problem. You should only use variable names and the following mathematical symbols: +, -, *, /, =.
The variable names should consist of letters and underscores.
You should not include any explanations or additional text in your response. Just provide the mathematical equation that represents the problem.
The equation should be in the form of a string, with variables and mathematical symbols only.
\end{lstlisting}

\subsection{Answer generation}

\begin{lstlisting}
{answer_generation_few_shot_examples}

Question: {natural language math problem}
Equations:
\end{lstlisting}

\subsection{Answer correction}

\begin{lstlisting}
{correction_one_shot_example}

Question: {natural language math problem}
Equations: 
{prior_response}

The equations have been extracted incorrectly for the following reasons:
{solvability}
{triggered_conditions}
Please extract the equations again for the given question.
{correction_suggestions}

Corrected Equations:
\end{lstlisting}

As mentioned in Section~\ref{sec:Methods} we used templates to generate the natural language explanation of the rules used. For Solvability we simply add the sentence ``The system is unsolvable'', while the templates for the triggered conditions and corresponding correction suggestions are reported in Table~\ref{tab:templates_explanations}.




\begin{table*}[h]
\centering
\caption{Summary of the templates for generating the explanation in natural language}
\label{tab:templates_explanations}
\resizebox{\linewidth}{!}{
\begin{tabular}{cll}
\toprule
\textbf{Type}          & \textbf{Triggered Conditions}        & \textbf{Correction Suggestions} \\
\midrule
Low  & The system may be under-specified
with too few equations & Add more equations to fully describe
the problem\\
High  & The system contains too many equations & Reduce the number of equations in the system\\
\midrule

Low  & Too few variables may miss
important aspects & Consider introducing additional
variables if needed\\
High  & Too many variables were used & Use fewer variables to simplify the equations\\
\midrule

Low  & Very few constants might miss
numeric details  & Include more relevant numerical values\\
High  & Too many constants were used  & Use fewer numeric constants \\
\midrule

Low  & Lack of addition operations
might indicate under-formed
expressions & Use addition operations where needed
to complete relationships\\
High  & Too many addition operations were used & Simplify the equations by reducing additions\\
\midrule

Low  & Too few multiplications might under-represent relationships & Include multiplication to model proportional or product-based relations\\
High  & Too many multiplication operations were used & Reduce the number of components in each equation \\
\midrule

Low  & Shallow equation depth might miss logical structure &Use more structured nesting to reflect problem hierarchy \\
High  & Equations are deeply nested & Simplify by reducing nesting in expressions\\
\midrule

Low  & Equations are overly simple & Add more structure to better represent the problem\\
High  & Equations are structurally complex with many elements & Reduce the number of components in each equation\\
\midrule

Low  & Too few leaf nodes may indicate underdeveloped equations & Use more complete expressions with relevant terms\\
High  & Too many terminal nodes in expression trees  & Reduce terminal terms for clarity \\
\midrule

Low  & The responses are overly uniform &Encourage more variation to explore diverse interpretations \\
High  & The responses are highly diverse  & Focus on extracting more consistent equations\\
\midrule

Low  & Very low variation detected & Consider encouraging alternative formulations \\
High  & There is significant variation among responses & Promote more consistent equation structures\\
\midrule

Low  & Strong consensus detected Still, double-check for correctness & Still, double-check for correctness despite agreement\\
High  & The majority answer is not clearly supported & Refine equations to better align with consensus \\
\bottomrule
            
\end{tabular}
}
\end{table*}

\newpage
\section{Datasets train-test split}\label{app:datasets}

This section details the experiments conducted to determine the optimar ratio for partitioning the datasets into training and testing sets. Given that our error detection method does not rely on large training sets, we explore different train-test splits to evaluate the impact of training set size on performance.
The results are reported in Table~\ref{table:split} and indicate that using a smaller training split yields better outcomes. This suggests that an effective rule generator can be learned from a small fraction of the available data, reducing computational costs while maintaining high performance.

\begin{table}[h]
\caption{Ablation Study Results on Train-Test Splits for DRAW-1k: The accuracy gain in error detection observed with different train-test split ratios for the DRAW-1k is shown. The goal is to identify a split that maximizes error detection performance while minimizing the training set size. The accuracy gain represents the gain in performance given by the error detection and correction and is computed as \tool ACC minus the initial answer generation ACC. We use Phi3 as answer generator and GPT4o as corrector.
}
\label{table:split}
\centering
\begin{tabular}{ccc}
\toprule
\textbf{Train/Test Split}	& \textbf{ACC Gain}\\
\midrule
0.1/0.9 &	   0.115\\
0.2/0.8 &	   0.111\\
0.3/0.7 &	   0.111\\
0.4/0.6 &	    0.109\\
0.5/0.5 &	    0.106\\
0.6/0.4 &	    0.108\\
0.7/0.3 &	    0.104\\
0.8/0.2 &	    0.104\\
0.9/0.1 &	     0.104\\ 
\bottomrule
\end{tabular}
\end{table}

\input{sections/ablations}

\section{Additional details}\label{sec:additional_details}

\subsection{Corrector accuracy analysis}
As shown in Table\ref{tab:corr_acc}, though a portion of correct examples are mistakenly flagged as incorrect, the overall initial accuracy of the detected samples is relatively low (31.2\%), and those samples achieve 80.9\% accuracy after correction. We also noticed that the True Positives, i.e., the samples incorrectly answered by the local LLM and correctly detected, are shown to be much more difficult and the stronger cloud-based LLM exhibits lower average accuracy on these examples, which is compatible with our intuition that our designed conditions are reflect the difficulty of the question.
\begin{table}[h]
\centering
\caption{Mean accuracy (ACC) across the examples detected as incorrect, also divided in True/False Positive. Experiment conducted on the DRAW-1k dataset, using Phi-3 for answer generation, GPT4o for error correction and $\varepsilon = 0.01$.}
\label{tab:corr_acc}
\begin{tabular}{lcc}
\toprule
\textbf{Subset} & \textbf{Initial ACC (\%)} & \textbf{Correction ACC (\%)} \\
\midrule
All Detected Examples              & 31.2   & 80.9 \\
False Positive  & 100   & 92.5 \\
True Positive & 0.0   & 75.7 \\
\bottomrule
\end{tabular}
\end{table}


\subsection{Full table~\ref{tab:results}}
Table~\ref{table1} extend the results reported in Table~\ref{tab:results}, including the number of LLM calls both for local and cloud-based queries.
\begin{table}[h!]
\caption{
Comparison of our method \tool with baselines (LLM only, SC and SC-solvable) and the state-of-the-art CRITIC~\cite{gou2023critic} framework using all considered LLMs (Phi-3, GPT4o, and DeepSeek) and optimal value of $\varepsilon$. Reported metrics include Accuracy (ACC) and number of calls to the LLM models separately for local and cloud-based. \tool achieves the best trade-off between these two metrics.\vspace{0.2cm}
}
\label{table1}
\centering
\begin{tabular}{lllcccccccc}\toprule 
 & & & \multicolumn{3}{c}{DRAW-1k} & &  \multicolumn{3}{c}{GSM-8k}
\\
\cmidrule{4-6} 
\cmidrule{8-10}
 &Answer& Error&   & \multicolumn{2}{c}{\# calls per sample}    &   & & \multicolumn{2}{c}{\# calls per sample} \\
Method &Generator&Corrector& ACC  & local & cloud  &  & ACC  & local & cloud \\
\midrule
LLM only & GPT4o & -& 91.6 & - & 1  & & 84.4 & - & 1 \\
LLM only & DeepSeek& - & 92.4 & - & 1  & & 86.2 & - & 1 \\
LLM only & Phi3 & - & 75.2 & 1 & - & & 52.2 & 1 & - \\
SC & Phi3 & - &  
78.8 & 10 & - &  & 
60.4 & 10 & -\\
SC+Solv.&Phi3&-&  
81.9 & 10 &  -&   & 
61.7 & 10 & -\\\midrule

\multirow{3}{*}[0pt]{CRITIC}&Phi3&Phi3&  
79.2 & 2 & -& &  
64 & 2 & -\\
&Phi3 & GPT4o&  
91.4& 1 & 1 &  & 
83.8 & 1& 1\\
&Phi3 & DeepSeek&  
92.7 & 1& 1 &  & 
84.6 & 1& 1\\

\midrule

\multirow{3}{*}[0pt]{\tool}&Phi3&Phi3&  
78.8 & 10.36& - &  & 
66.2 & 10.43 & -\\
&Phi3 &GPT4o&  
85.7 & 10 & 0.36 &  & 
74.4 & 10& 0.43\\
&Phi3 &DeepSeek&  
87.8 & 10 & 0.36  & &
75.0 & 10& 0.43\\

\bottomrule
\end{tabular}
\end{table}

\subsection{Additional details on CRITIC setup}

While the original CRITIC pipeline used the same LLM model for both initial answer generation and correction stages, our pipeline utilizes different LLM models for these two phases. For fair comparison, we adapted the original CRITIC pipeline to use two different models as well.

When using a smaller local model for initial generation and a stronger cloud-based LLM for correction, we observed that a single correction loop in the CRITIC pipeline already achieves performance comparable to the cloud-based LLM-only baseline.

Unlike the original pipeline, where additional correction loops resulted in higher final performance, and they use 4-loops as the default setting, our implementation showed no significant performance gains from multiple loops. But increasing the number of loops from 1 to 4 means increasing the correction re-prompt rate from 100\% to 400\%. Therefore, we set the re-prompt rate of the CRITIC pipeline to only 100\% in our experiment.

The original CRITIC approach was designed for a different output syntax than our implementation, which is to generate a executable Python code for math word problems, where our approach focuses on generating equation systems and then utilizing an external tool (SymPy) to solve them. For fair comparison, we modified CRITIC's implementation to align with equation generation.



\section{Robustness of Equation Distance Analysis}
\label{sec:equation-distance-analysis}
This section validates the conclusion from Section~\ref{subsec:sol-improvement-analysis}--that EDCIM's correction improves solution quality--by testing the robustness of this finding. First, we provide a more granular analysis of the primary distance metric used in the main paper. Second, we explore several alternative methodologies for generating equation embeddings to ensure our findings are not dependent on a single measurement technique. The collective results demonstrate that while the magnitude of improvement can vary, the overall positive impact of our correction framework is a consistent and fundamental aspect of its design.

\subsection{Detailed Analysis of the Primary Distance Metric}
\begin{figure}[H]
    \centering
    \includegraphics[width=0.6\linewidth]{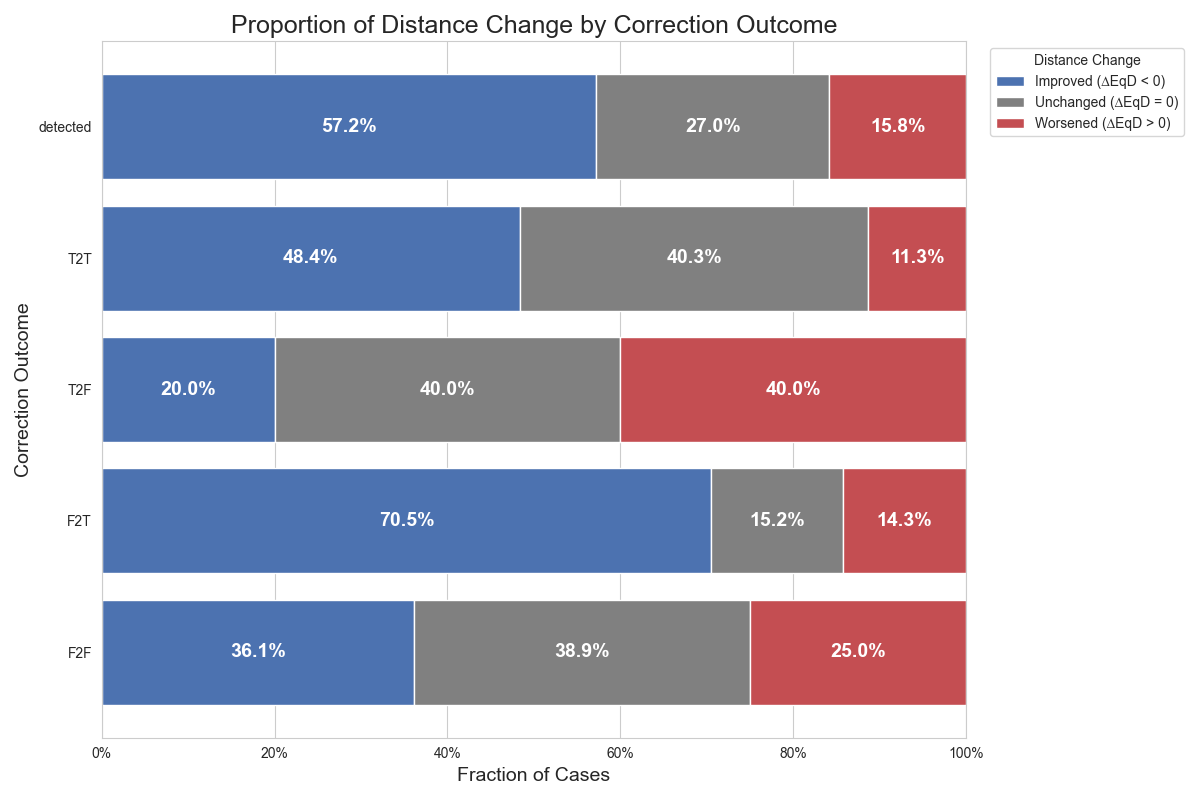}
    \caption{A detailed bar plot analysis of the primary distance metric used in the main paper, showing the percentage of outcomes that improved, remained unchanged, or degraded.}
    \label{fig:main-metric-bars}
\end{figure}

While the violin plot in Figure~\ref{fig:violin}
provides a high-level view of the distribution, a bar plot offers a more granular perspective by explicitly quantifying the proportion of solutions that improved (negative $\Delta EqD$), degraded (positive $\Delta EqD$), or remained unchanged. As shown in Figure~\ref{fig:main-metric-bars}, this metric confirms the key trends. Successful corrections (\textbf{F2T}) are overwhelmingly positive, with over 70\% of cases improving. Degradations of correct solutions (\textbf{T2F}) are also clear. For the challenging False-to-False (\textbf{F2F}) cases, this specific metric shows that the solution quality is more often improved or unchanged than degraded, supporting the value of the correction step even when a perfect solution is not reached.

\subsection{Robustness Analysis with Alternative Methodologies}

\subsubsection{Alternative Embedding Methodologies}
To test the robustness of our findings, we created alternative vector representations for each system of equations using the following four methods:
\begin{itemize}
    \item \textbf{Node2Vec-based Embeddings:} Each system of equations was modeled as a graph, with variables and operators as nodes. We then utilized the node2vec algorithm to learn a low-dimensional vector embedding that captures the graph's topological structure.
    \item \textbf{GNN-based Embeddings:} To leverage learned structural features, we utilized a Graph Neural Network (GNN) pre-trained on a graph classification task using the MathQA dataset. This pre-training process enables the GNN to generate embeddings that capture the semantic and structural nuances inherent in mathematical reasoning problems. For our analysis, we passed each equation graph through this pre-trained GNN and extracted the final aggregated graph-level hidden state.
    \item \textbf{LLM-based Embeddings:} We explored two domain-specialized language models: MathBERT and GraphCodeBERT-base. Each equation system was serialized into a structured string, tokenized, and input into each model. The output embedding corresponding to the `[CLS]` token was extracted from the final hidden layer to serve as the contextual vector representation.
    \item \textbf{TransformersConv-based Embeddings:} We employed a Transformer-based graph convolutional layer (TransformersConv). This method uses a multi-head self-attention mechanism, allowing each node to dynamically weigh the importance of its neighbors when updating its representation. The final graph embedding is produced by pooling the node-level embeddings.
\end{itemize}

\subsection{Distance Metrics and Results}
For each embedding methodology, we calculated the change in distance $(\Delta EqD)$ using four standard metrics (Cosine, Euclidean, Manhattan, and Correlation Distance). To demonstrate our findings without duplicating all results, we present a representative sample from this analysis.

The Figura~\ref{fig:node2vec-manhattan} and Figure~\ref{fig:llm-manhattan} below show the results using the \textbf{Manhattan} distance for the Node2vec and LLM-based embeddings. This metric was chosen as it was consistently illustrative across different methods. A negative $(\Delta EqD)$ indicates that the corrected equation system is closer to the ground truth than the initial one.

\begin{figure}[H]
    \centering
    \includegraphics[width=0.7\linewidth]{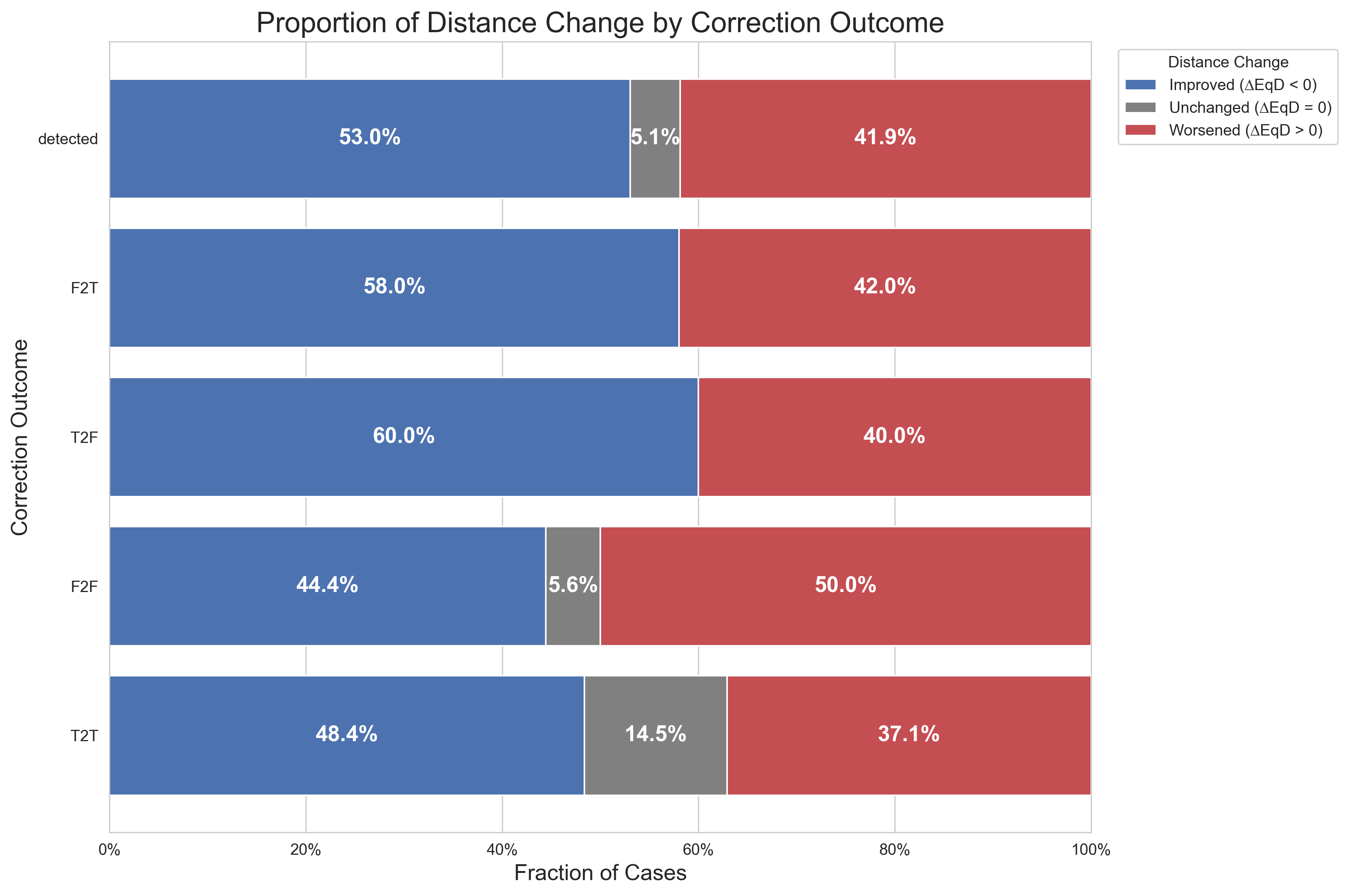}
    \caption{Results using Node2vec Embeddings with Manhattan Distance.}
    \label{fig:node2vec-manhattan}
\end{figure}

\begin{figure}[H]
    \centering
    \includegraphics[width=0.7\linewidth]{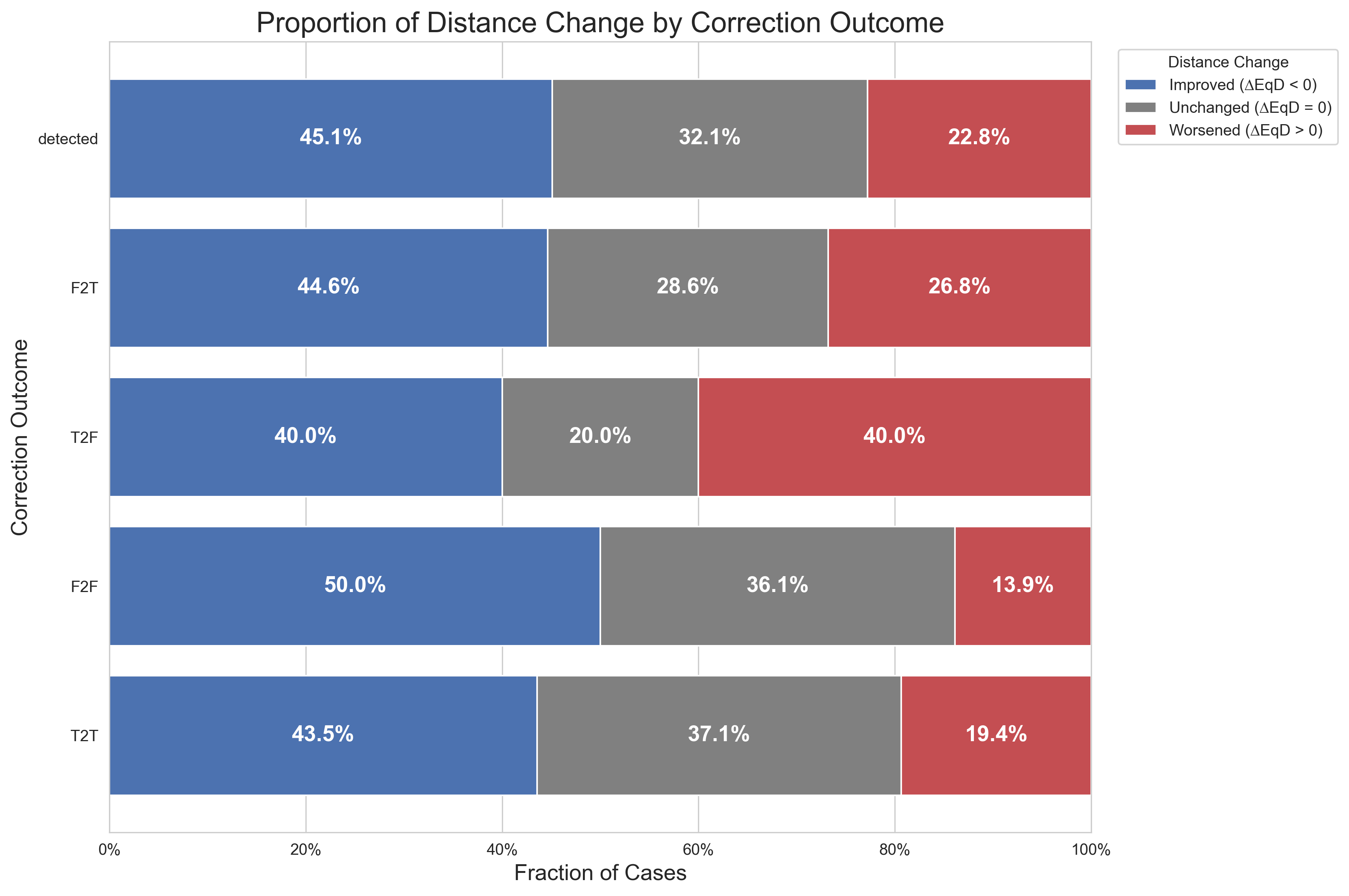}
    \caption{Results using LLM Embeddings with Manhattan Distance.}
    \label{fig:llm-manhattan}
\end{figure}

\subsection{Summary of Findings}
The analysis presented in this appendix supports the findings reported in Section~\ref{subsec:sol-improvement-analysis}. The detailed analysis of our primary metric (Figure~\ref{fig:main-metric-bars}) confirms its ability to capture meaningful improvements. Furthermore, the exploration of alternative embedding and distance methodologies, sampled in Figures~\ref{fig:node2vec-manhattan} and \ref{fig:llm-manhattan}, reinforces the main conclusion: EDCIM's correction mechanism provides a net positive impact on solution quality. While the specific effect on False-to-False cases varies across metrics, the framework's ability to consistently and significantly improve solutions in False-to-True scenarios validates its overall effectiveness.







%% file: sections/ablations.tex
\section{Ablation studies}
\label{sec:Ablation}
In this section, we present a series of ablation studies to evaluate the impact of different parts of our framework. In particular, we experiment with various types of detectors (including EDR, oracle, and total re-prompting), rule types, and prompt designs. Additional results and analysis can be found in the Appendix.

\subsection{Impact of: detector type}
\begin{table*}[h]
\caption{Detection performance (F1-score and re-prompt rate) of five error detector models and their impact on the correctness (ACC) of the final results for \tool and CRITIC, evaluated on the DRAW-1K and GSM-8K datasets. Results are based on Phi-3 as the answer generator and GPT4o as the error corrector.}
\label{table2}
\centering
\small
\begin{tabular}{lcccccccc}\toprule
\multirow{4}{*}[0pt]{Error Detector}&\multicolumn{4}{c}{DRAW-1k}&\multicolumn{4}{c}{GSM-8k}\\\cmidrule(lr){2-5}\cmidrule(lr){6-9}
&\multicolumn{2}{c}{Detector}&\multicolumn{2}{c}{ACC}&\multicolumn{2}{c}{Detector}&\multicolumn{2}{c}{ACC}\\
\cmidrule(lr){2-3}\cmidrule(lr){4-5}\cmidrule(lr){6-7}\cmidrule(lr){8-9}
& re-prompt & f1 & CRITIC & \tool & re-prompt & f1 & CRITIC & \tool \\\midrule
All\tnote{1}&1&0.4&91.4&91.6&1&0.65&83.8&84.4\\
Oracle\tnote{2}&0.248&1&92.2&93.4&0.48&1&86.6&86.4\\
Solvability\tnote{3}&0.13&0.71&86.6&86.1&0.21&0.64&69.2&69.5\\
EDR w/o Solv.\tnote{4}&0.28&0.36&83.3&83.5&0.31&0.67&69.7&70.2\\
EDR + Solv. (ours)\tnote{5}&0.36&0.59&85.2&85.7&0.43&0.71&73.7&74.4\\
\bottomrule

\end{tabular}
\end{table*}

Table \ref{table2} evaluates the impact of combining different error detector models on the correction accuracy. 
We considered 5 different error detector models:
(1) \textbf{All} assume all samples are incorrect and re-prompt all of them. This corresponds to CRITIC detector.
(2) \textbf{Oracle} use ground truth systems of equations to trigger correction only on the incorrect answers.
(3) \textbf{Solvability} incorporates feedback from SymPy to assess whether the generated equations are solvable. If the equations are not solvable, triggers the error correction process, including this information in the prompt.
(4) \textbf{EDR w/o Solvability} use EDR detection without checking if the equations are solvable.
(5) \textbf{EDR + Solvablity (\tool)} combines EDR detection with a solvability check. Both types of feedback, from EDR and the equation solvability test, are included in the correction prompt. This represents the standard configuration for \tool.

The EDR detector alone provides strong detection but may miss solvability issues, while the solvability criterion alone filters unsolvable equations but lacks full error detection capabilities. When combined, the re-prompt rate increases slightly, but precision and overall detection accuracy significantly improve by reducing false negatives and enhancing true error identification. This demonstrates the advantage of integrating multiple detection mechanisms for more reliable corrections.

\subsection{Impact of: prompt types}




Table~\ref{table3} shows the impact of different prompt components affect \tool performance in terms of accuracy (ACC). We modified the prompt by removing components (see Section \ref{sec:set-up}) by considering the 7 settings provided in the table. The first setting correspond to the standard method employed in \tool, while the last setting correspond to a static prompt that does not take into account dynamic, example-dependent cues in guiding responses.

\begin{table}[h]
\caption{Impact of different prompt settings on \tool performance (ACC) for the DRAW-1K and GSM-8K datasets. Results are based on Phi-3 as answer generator and GPT4o as error corrector.}
\label{table3}
\centering
\begin{tabular}{ccccccc}\toprule
& \multicolumn{4}{c}{prompt setting}&\multicolumn{2}{c}{\tool ACC }\\
\cmidrule(lr){2-5}
\cmidrule(lr){6-7}
setting & few-shots & solvability & violations & suggestions & DRAW-1K & GSM-8K\\
\midrule
1& \checkmark&\checkmark&\checkmark&\checkmark&85.7&74.4\\
2& -&\checkmark&\checkmark&\checkmark&81.8&70.3\\
3& \checkmark&-&\checkmark&\checkmark&84.2&73.9\\
4& \checkmark&\checkmark&-&\checkmark&84.5&74.1\\
5& \checkmark&-&-&\checkmark&84.2&73.7\\
6& \checkmark&\checkmark&\checkmark&-&84.5&73.9\\
7& \checkmark&-&-&-&83.7&73.2\\
\bottomrule
\end{tabular}
\end{table}

The results indicate that the standard setting used in \tool (setting 1) provides the highest overall performance.
We conjecture that violated conditions and correction suggestions lead to very similar additional information: violated conditions identify what went wrong, while correction suggestions offer guidance on how to resolve these errors.
However, in the GSM-8K dataset, using correction suggestions instead of violated conditions yields slightly better performance, suggesting that more explicit guidance can be beneficial in certain cases (see the comparison between settings 5 and 7).
We also note that the static prompt, which includes no example-dependent information, leads to the worst performance overall, highlighting the importance of targeted example-dependent feedback.

\subsection{Impact of: rules used by our EDR detector}

\begin{table}[h]
\caption{Impact of different meta-rule categories on EDR detector performance on DRAW-1K, using Phi-3 for answer generation, GPT4o for error correction and $\varepsilon = $
0.06
}
\label{table4}
\centering
\begin{tabular}{cccccccc}\toprule

\multicolumn{3}{c}{Rule Setting}&\multicolumn{3}{c}{EDR Detector performance} & \multicolumn{2}{c}{\tool} \\ 
\cmidrule(lr){1-3}\cmidrule(lr){4-6}\cmidrule(lr){7-8}

{\#Symbols}&{Diversity}&{Complexity}&{Precision}&{Recall}&{f1}&{re-prompt \%}&{ACC}\\\midrule
\checkmark&\checkmark&\checkmark&48.8&74.7&59.0&38.3& 87.7 \\ 
\checkmark&\checkmark&-&47.9&73.1&57.9&38.2& 87.5 \\ 
\checkmark&-&\checkmark&56.8&70.3&62.8&31.0& 87.3 \\ 
\checkmark&-&-&56.8&70.3&62.8&31.0& 87.3 \\ 
-&\checkmark&\checkmark&60.8&65.5&63.1&26.9& 87.0 \\ 
-&\checkmark&-&62.8&63.1&62.9&25.1& 86.6 \\ 
-&-&\checkmark&76.1&60.2&67.3&19.8& 86.0 \\ 
\bottomrule
\end{tabular}
\end{table}

Table~\ref{table4} shows how EDR detector performance changes when using different combinations of our three designed meta-rule groups. We find that applying more rule types causes the detector to flag more samples as errors, resulting in decreased precision but higher re-prompt rates. 
Despite the lower precision, this increasing overall accuracy. 

%% file: main.bbl
\begin{thebibliography}{34}
\providecommand{\natexlab}[1]{#1}

\bibitem[{Abdin et~al.(2024)Abdin, Aneja, Awadalla, Awadallah, Awan, Bach,
  Bahree, Bakhtiari, Bao, Behl et~al.}]{abdin2024phi}
Abdin, M.; Aneja, J.; Awadalla, H.; Awadallah, A.; Awan, A.~A.; Bach, N.;
  Bahree, A.; Bakhtiari, A.; Bao, J.; Behl, H.; et~al. 2024.
\newblock Phi-3 technical report: A highly capable language model locally on
  your phone.
\newblock \emph{arXiv preprint arXiv:2404.14219}.

\bibitem[{Achiam et~al.(2023)Achiam, Adler, Agarwal, Ahmad, Akkaya, Aleman,
  Almeida, Altenschmidt, Altman, Anadkat et~al.}]{achiam2023gpt}
Achiam, J.; Adler, S.; Agarwal, S.; Ahmad, L.; Akkaya, I.; Aleman, F.~L.;
  Almeida, D.; Altenschmidt, J.; Altman, S.; Anadkat, S.; et~al. 2023.
\newblock Gpt-4 technical report.
\newblock \emph{arXiv preprint arXiv:2303.08774}.

\bibitem[{An et~al.(2023)An, Ma, Lin, Zheng, Lou, and Chen}]{an2023learning}
An, S.; Ma, Z.; Lin, Z.; Zheng, N.; Lou, J.-G.; and Chen, W. 2023.
\newblock Learning from mistakes makes llm better reasoner.
\newblock \emph{arXiv preprint arXiv:2310.20689}.

\bibitem[{Cobbe et~al.(2021)Cobbe, Kosaraju, Bavarian, Chen, Jun, Kaiser,
  Plappert, Tworek, Hilton, Nakano, Hesse, and Schulman}]{cobbe2021gsm8k}
Cobbe, K.; Kosaraju, V.; Bavarian, M.; Chen, M.; Jun, H.; Kaiser, L.; Plappert,
  M.; Tworek, J.; Hilton, J.; Nakano, R.; Hesse, C.; and Schulman, J. 2021.
\newblock Training Verifiers to Solve Math Word Problems.
\newblock \emph{arXiv preprint arXiv:2110.14168}.

\bibitem[{Cornelio et~al.(2023)Cornelio, Stuehmer, Hu, and
  Hospedales}]{cornelio_2023_NASR}
Cornelio, C.; Stuehmer, J.; Hu, S.~X.; and Hospedales, T. 2023.
\newblock Learning where and when to reason in neuro-symbolic inference.
\newblock In \emph{The Eleventh International Conference on Learning
  Representations (ICLR)}.

\bibitem[{Didolkar et~al.(2024)Didolkar, Goyal, Ke, Guo, Valko, Lillicrap,
  Jimenez~Rezende, Bengio, Mozer, and Arora}]{didolkar2024metacognitive}
Didolkar, A.; Goyal, A.; Ke, N.~R.; Guo, S.; Valko, M.; Lillicrap, T.;
  Jimenez~Rezende, D.; Bengio, Y.; Mozer, M.~C.; and Arora, S. 2024.
\newblock Metacognitive capabilities of llms: An exploration in mathematical
  problem solving.
\newblock \emph{Advances in Neural Information Processing Systems}, 37:
  19783--19812.

\bibitem[{Flavell(1976)}]{flavell1976metacognitive}
Flavell, J.~H. 1976.
\newblock Metacognitive aspects of problem solving.
\newblock In \emph{The nature of intelligence}, 231--236. Routledge.

\bibitem[{Friel and Sanyal(2023)}]{friel2023chainpoll}
Friel, R.; and Sanyal, A. 2023.
\newblock Chainpoll: A high efficacy method for llm hallucination detection.
\newblock \emph{arXiv preprint arXiv:2310.18344}.

\bibitem[{Gilpin et~al.(2018)Gilpin, Bau, Yuan, Bajwa, Specter, and
  Kagal}]{gilpin2018explaining}
Gilpin, L.~H.; Bau, D.; Yuan, B.~Z.; Bajwa, A.; Specter, M.; and Kagal, L.
  2018.
\newblock Explaining explanations: An overview of interpretability of machine
  learning.
\newblock In \emph{2018 IEEE 5th International Conference on data science and
  advanced analytics (DSAA)}, 80--89. IEEE.

\bibitem[{Gou et~al.(2023)Gou, Shao, Gong, Shen, Yang, Duan, and
  Chen}]{gou2023critic}
Gou, Z.; Shao, Z.; Gong, Y.; Shen, Y.; Yang, Y.; Duan, N.; and Chen, W. 2023.
\newblock Critic: Large language models can self-correct with tool-interactive
  critiquing.
\newblock \emph{arXiv preprint arXiv:2305.11738}.

\bibitem[{He-Yueya et~al.(2023)He-Yueya, Poesia, Wang, and
  Goodman}]{he2023solving}
He-Yueya, J.; Poesia, G.; Wang, R.~E.; and Goodman, N.~D. 2023.
\newblock Solving math word problems by combining language models with symbolic
  solvers.
\newblock \emph{arXiv preprint arXiv:2304.09102}.

\bibitem[{Huang et~al.(2023)Huang, Yu, Ma, Zhong, Feng, Wang, Chen, Peng, Feng,
  Qin et~al.}]{huang2023survey}
Huang, L.; Yu, W.; Ma, W.; Zhong, W.; Feng, Z.; Wang, H.; Chen, Q.; Peng, W.;
  Feng, X.; Qin, B.; et~al. 2023.
\newblock A survey on hallucination in large language models: Principles,
  taxonomy, challenges, and open questions, 2023.
\newblock \emph{arXiv preprint arXiv:2311.05232}.

\bibitem[{Imani, Du, and Shrivastava(2023)}]{imani2023mathprompter}
Imani, S.; Du, L.; and Shrivastava, H. 2023.
\newblock Mathprompter: Mathematical reasoning using large language models.
\newblock \emph{arXiv preprint arXiv:2303.05398}.

\bibitem[{Kamoi et~al.(2024)Kamoi, Zhang, Zhang, Han, and Zhang}]{kamoi2024can}
Kamoi, R.; Zhang, Y.; Zhang, N.; Han, J.; and Zhang, R. 2024.
\newblock When can llms actually correct their own mistakes? a critical survey
  of self-correction of llms.
\newblock \emph{Transactions of the Association for Computational Linguistics},
  12: 1417--1440.

\bibitem[{Kricheli et~al.(2024)Kricheli, Vo, Datta, Ozgur, and
  Shakarian}]{kricheli2024error}
Kricheli, J.~S.; Vo, K.; Datta, A.; Ozgur, S.; and Shakarian, P. 2024.
\newblock Error detection and constraint recovery in hierarchical multi-label
  classification without prior knowledge.
\newblock In \emph{Proceedings of the 33rd ACM International Conference on
  Information and Knowledge Management}, 3842--3846.

\bibitem[{Lee et~al.(2024)Lee, Ngu, Sahdev, Motaganahall, Chowdhury, Xi, and
  Shakarian}]{lee2024metal}
Lee, N.; Ngu, N.; Sahdev, H.~S.; Motaganahall, P.; Chowdhury, A. M.~S.; Xi, B.;
  and Shakarian, P. 2024.
\newblock Metal Price Spike Prediction via a Neurosymbolic Ensemble Approach.
\newblock \emph{arXiv preprint arXiv:2410.12785}.

\bibitem[{Lewkowycz et~al.(2022)Lewkowycz, Andreassen, Dohan, Dyer,
  Michalewski, Ramasesh, Slone, Anil, Schlag, Gutman-Solo
  et~al.}]{lewkowycz2022solving}
Lewkowycz, A.; Andreassen, A.; Dohan, D.; Dyer, E.; Michalewski, H.; Ramasesh,
  V.; Slone, A.; Anil, C.; Schlag, I.; Gutman-Solo, T.; et~al. 2022.
\newblock Solving quantitative reasoning problems with language models.
\newblock \emph{Advances in Neural Information Processing Systems}, 35:
  3843--3857.

\bibitem[{Liu et~al.(2024)Liu, Feng, Xue, Wang, Wu, Lu, Zhao, Deng, Zhang, Ruan
  et~al.}]{liu2024deepseek}
Liu, A.; Feng, B.; Xue, B.; Wang, B.; Wu, B.; Lu, C.; Zhao, C.; Deng, C.;
  Zhang, C.; Ruan, C.; et~al. 2024.
\newblock Deepseek-v3 technical report.
\newblock \emph{arXiv preprint arXiv:2412.19437}.

\bibitem[{Madaan et~al.(2023)Madaan, Tandon, Gupta, Hallinan, Gao, Wiegreffe,
  Alon, Dziri, Prabhumoye, Yang et~al.}]{madaan2023self}
Madaan, A.; Tandon, N.; Gupta, P.; Hallinan, S.; Gao, L.; Wiegreffe, S.; Alon,
  U.; Dziri, N.; Prabhumoye, S.; Yang, Y.; et~al. 2023.
\newblock Self-refine: Iterative refinement with self-feedback.
\newblock \emph{Advances in Neural Information Processing Systems}, 36:
  46534--46594.

\bibitem[{Meurer et~al.(2017)Meurer, Smith, Paprocki, \v{C}ert\'{i}k,
  Kirpichev, Rocklin, Kumar, Ivanov, Moore, Singh, Rathnayake, Vig, Granger,
  Muller, Bonazzi, Gupta, Vats, Johansson, Pedregosa, Curry, Terrel,
  Rou\v{c}ka, Saboo, Fernando, Kulal, Cimrman, and Scopatz}]{sympy}
Meurer, A.; Smith, C.~P.; Paprocki, M.; \v{C}ert\'{i}k, O.; Kirpichev, S.~B.;
  Rocklin, M.; Kumar, A.; Ivanov, S.; Moore, J.~K.; Singh, S.; Rathnayake, T.;
  Vig, S.; Granger, B.~E.; Muller, R.~P.; Bonazzi, F.; Gupta, H.; Vats, S.;
  Johansson, F.; Pedregosa, F.; Curry, M.~J.; Terrel, A.~R.; Rou\v{c}ka, v.;
  Saboo, A.; Fernando, I.; Kulal, S.; Cimrman, R.; and Scopatz, A. 2017.
\newblock SymPy: symbolic computing in Python.
\newblock \emph{PeerJ Computer Science}, 3: e103.

\bibitem[{Mishra et~al.(2024)Mishra, Soliman, Ramakrishna, Galstyan, and
  Kumar}]{mishra2024correcting}
Mishra, K.; Soliman, T.; Ramakrishna, A.; Galstyan, A.; and Kumar, A. 2024.
\newblock Correcting Language Model Outputs by Editing Salient Layers.
\newblock In \emph{Findings of the Association for Computational Linguistics:
  EACL 2024}, 1295--1305.

\bibitem[{Ngu, Lee, and Shakarian(2024)}]{ngu2024diversity}
Ngu, N.; Lee, N.; and Shakarian, P. 2024.
\newblock Diversity measures: Domain-independent proxies for failure in
  language model queries.
\newblock In \emph{2024 IEEE 18th International Conference on Semantic
  Computing (ICSC)}, 176--182. IEEE.

\bibitem[{Pan et~al.(2023{\natexlab{a}})Pan, Albalak, Wang, and
  Wang}]{pan2023logic}
Pan, L.; Albalak, A.; Wang, X.; and Wang, W.~Y. 2023{\natexlab{a}}.
\newblock Logic-lm: Empowering large language models with symbolic solvers for
  faithful logical reasoning.
\newblock \emph{arXiv preprint arXiv:2305.12295}.

\bibitem[{Pan et~al.(2023{\natexlab{b}})Pan, Saxon, Xu, Nathani, Wang, and
  Wang}]{pan2023automatically}
Pan, L.; Saxon, M.; Xu, W.; Nathani, D.; Wang, X.; and Wang, W.~Y.
  2023{\natexlab{b}}.
\newblock Automatically correcting large language models: Surveying the
  landscape of diverse self-correction strategies.
\newblock \emph{arXiv preprint arXiv:2308.03188}.

\bibitem[{Paul et~al.(2023)Paul, Ismayilzada, Peyrard, Borges, Bosselut, West,
  and Faltings}]{paul2023refiner}
Paul, D.; Ismayilzada, M.; Peyrard, M.; Borges, B.; Bosselut, A.; West, R.; and
  Faltings, B. 2023.
\newblock Refiner: Reasoning feedback on intermediate representations.
\newblock \emph{arXiv preprint arXiv:2304.01904}.

\bibitem[{Shakarian et~al.(2023)Shakarian, Koyyalamudi, Ngu, and
  Mareedu}]{shakarian2023independent}
Shakarian, P.; Koyyalamudi, A.; Ngu, N.; and Mareedu, L. 2023.
\newblock An independent evaluation of ChatGPT on mathematical word problems
  (MWP).
\newblock \emph{arXiv preprint arXiv:2302.13814}.

\bibitem[{Shakarian, Simari, and Bastian(2025)}]{shakarian2025probabilistic}
Shakarian, P.; Simari, G.~I.; and Bastian, N.~D. 2025.
\newblock Probabilistic Foundations for Metacognition via Hybrid-AI.
\newblock \emph{arXiv preprint arXiv:2502.05398}.

\bibitem[{Su et~al.(2024)Su, Wang, Ai, Hu, Wu, Zhou, and
  Liu}]{su2024unsupervised}
Su, W.; Wang, C.; Ai, Q.; Hu, Y.; Wu, Z.; Zhou, Y.; and Liu, Y. 2024.
\newblock Unsupervised real-time hallucination detection based on the internal
  states of large language models.
\newblock \emph{arXiv preprint arXiv:2403.06448}.

\bibitem[{Upadhyay and Chang(2016)}]{upadhyay2016annotating}
Upadhyay, S.; and Chang, M.-W. 2016.
\newblock Annotating derivations: A new evaluation strategy and dataset for
  algebra word problems.
\newblock \emph{arXiv preprint arXiv:1609.07197}.

\bibitem[{Upadhyaya and Sridharamurthy(2024)}]{upadhyaya2024internalized}
Upadhyaya, N.; and Sridharamurthy, R. 2024.
\newblock Internalized Self-Correction for Large Language Models.
\newblock \emph{arXiv preprint arXiv:2412.16653}.

\bibitem[{Xi et~al.(2023)Xi, Scaria, Bavikadi, and Shakarian}]{xi2023rule}
Xi, B.; Scaria, K.; Bavikadi, D.; and Shakarian, P. 2023.
\newblock Rule-based error detection and correction to operationalize movement
  trajectory classification.
\newblock \emph{arXiv preprint arXiv:2308.14250}.

\bibitem[{Yamauchi et~al.(2023)Yamauchi, Sonoda, Sannai, and
  Kumagai}]{yamauchi2023lpml}
Yamauchi, R.; Sonoda, S.; Sannai, A.; and Kumagai, W. 2023.
\newblock LPML: llm-prompting markup language for mathematical reasoning.
\newblock \emph{arXiv preprint arXiv:2309.13078}.

\bibitem[{Ye et~al.(2023)Ye, Chen, Dillig, and Durrett}]{ye2023satlm}
Ye, X.; Chen, Q.; Dillig, I.; and Durrett, G. 2023.
\newblock Satlm: Satisfiability-aided language models using declarative
  prompting.
\newblock \emph{Advances in Neural Information Processing Systems}, 36:
  45548--45580.

\bibitem[{Zhao et~al.(2023)Zhao, Zhang, Si, Nan, Tang, and
  Cohan}]{zhao2023investigating}
Zhao, Y.; Zhang, H.; Si, S.; Nan, L.; Tang, X.; and Cohan, A. 2023.
\newblock Investigating Table-to-Text Generation Capabilities of LLMs in
  Real-World Information Seeking Scenarios.
\newblock \emph{arXiv preprint arXiv:2305.14987}.

\end{thebibliography}
